\pdfoutput=1

\documentclass[11pt]{article}

\usepackage[preprint]{acl}

\usepackage{times}
\usepackage{latexsym}
\usepackage{lipsum}  
\usepackage{booktabs}
\usepackage{adjustbox}
\usepackage{url}
\usepackage{amssymb}
\usepackage{amsmath}
\usepackage{multirow}
\usepackage{tabularx}
\usepackage{graphicx}
\usepackage{enumitem}
\usepackage{multirow}
\usepackage{mathtools}
\usepackage{tcolorbox}

\usepackage{xcolor}

\definecolor{mydarkgreen}{RGB}{20, 170,20}

\newcommand{\dataset}{\textsc{MissciPlus}}
\newcommand{\missci}{\textsc{Missci}}

\newcommand{\maybeadd}[1]{}

\usepackage[T1]{fontenc}

\usepackage[utf8]{inputenc}

\usepackage{microtype}

\usepackage{inconsolata}

\usepackage{graphicx}
\usepackage{todonotes}

\title{Grounding Fallacies Misrepresenting Scientific Publications in Evidence}

\author{
 Max Glockner$^\clubsuit$,
 Yufang Hou$^{\diamondsuit \heartsuit}$,
 Preslav Nakov$^\spadesuit$ \and
 Iryna Gurevych$^{\clubsuit}$
 \vspace{0.25em}\\
  $^\clubsuit$Ubiquitous Knowledge Processing Lab (UKP Lab),\\
TU Darmstadt and Hessian Center for AI (hessian.AI)\\
  $^\diamondsuit$IT:U Interdisciplinary Transformation University Austria \\ $^\heartsuit$IBM Research Ireland,
  $^\spadesuit$MBZUAI
  \\
    \url{www.ukp.tu-darmstadt.de}
}

\begin{document}
\maketitle
\begin{abstract}
Health-related misinformation claims often falsely cite a credible biomedical publication as evidence. %
These publications only superficially seem to support the false claim, when logical fallacies are applied.
In this work, we aim to detect and to highlight such fallacies, which requires assessing the exact content of the misrepresented publications. To achieve this, we introduce \dataset{}, an extension of the fallacy detection dataset \missci{}.
\dataset{} extends \missci{} by grounding the applied fallacies in real-world passages from misrepresented studies. This creates a realistic test-bed for detecting and verbalizing fallacies under real-world input conditions, and enables new and realistic passage-retrieval tasks.
\dataset{} is the first logical fallacy dataset which pairs the real-world misrepresented evidence with incorrect claims, identical to the input to evidence-based fact-checking models. With \dataset{}, we \emph{i}) benchmark retrieval models in identifying passages that support claims only with fallacious reasoning, \emph{ii)} evaluate how well LLMs verbalize fallacious reasoning based on misrepresented scientific passages, and \emph{iii)} assess the effectiveness of fact-checking models in refuting claims that misrepresent biomedical research.
Our findings show that current fact-checking models struggle to use misrepresented scientific passages to refute misinformation. Moreover, these passages can mislead LLMs into accepting false claims as true.\footnote{Code and data are available at \url{https://github.com/UKPLab/naacl2025-missciplus}}

\end{abstract}

\section{Introduction}
\label{sec:intro}
\begin{figure}
\small
    \centering
    \includegraphics[width=\linewidth]{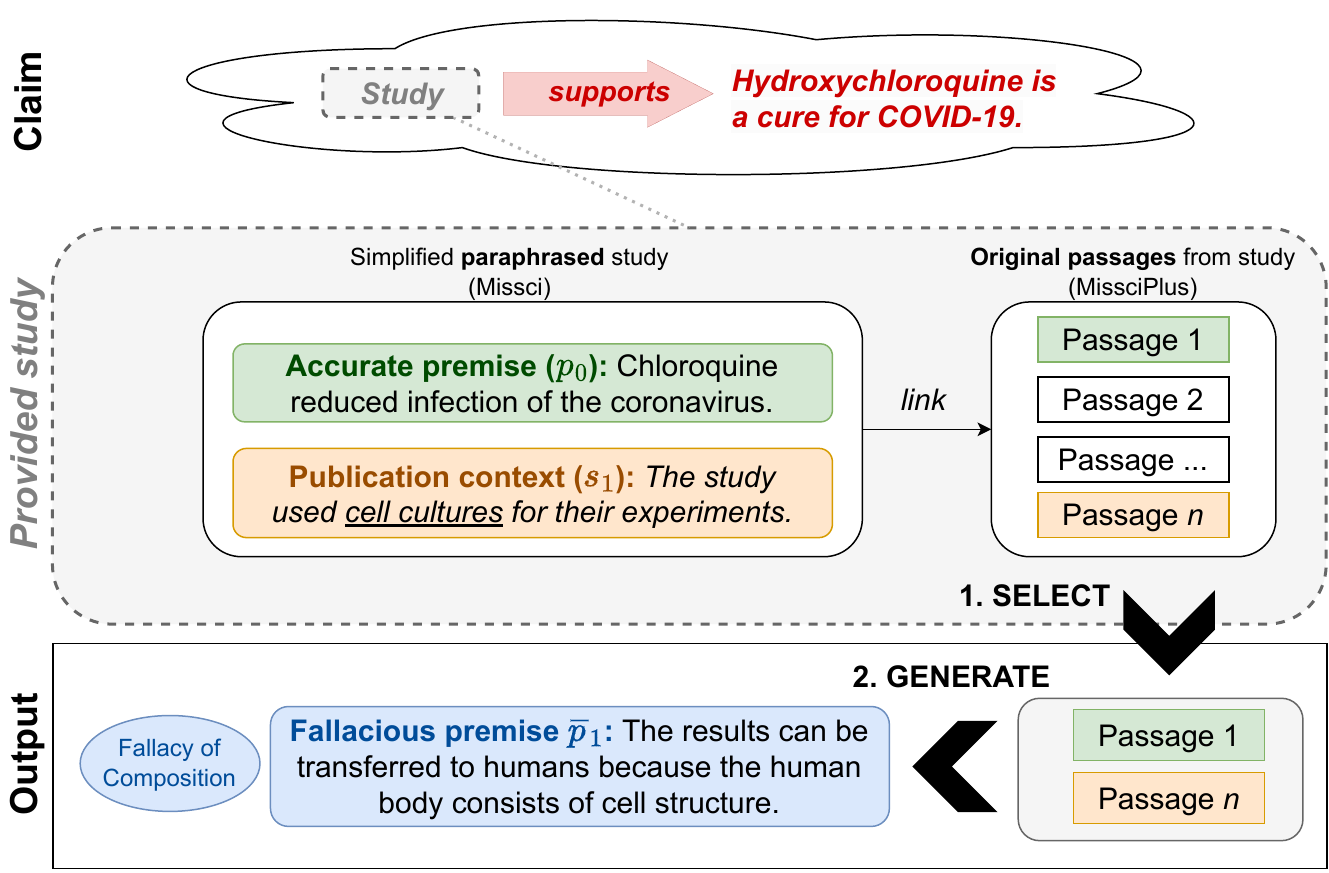}
    \caption{ 
    We link the paraphrased context from \missci{} to real-world passages. The LLM must (\emph{i})~find relevant passages from the original study and (\emph{ii})~generate a fallacious premise to (falsely) support the claim.
    }
    \label{fig:figure-main}
\end{figure}

Health-related misinformation has caused significant harm in our society \citep{zarocostas2020fight}. Human fact-checking (HFC), which is time-consuming, must prioritize the most impactful claims and struggles with the rapid spread of misinformation \citep{arnold2020challenges, vosoughi2018spread}. Two main approaches can automatically combat health-related misinformation: Scientific automated fact-checking (AFC) retrieves scientific evidence documents to support or to refute a claim, but faces challenges with mismatched specificity between claim and evidence \citep{wadden-etal-2022-scifact}, and the reliance on the availability of refuting evidence \citep{glockner-etal-2022-missing}. Reason-checking \citep{musi2023developing} rejects claims that use fallacious reasoning, 
which is particularly relevant %
since facts are often misrepresented or skewed \citep{brennen2020types}.
Unlike most logical fallacy detection datasets, which assume that fallacious reasoning is explicitly stated \citep{jin-etal-2022-logical, alhindi-etal-2022-multitask}, 
real-world scenarios often involve fallacies that are not explicitly articulated.

The \missci{} dataset \citep{glockner2024dismantling} addresses implicit fallacious reasoning by reconstructing the \emph{fallacious logical argument}, explicitly verbalizing the fallacies that led to the incorrect claim.  However, \missci{} provides pre-selected and simplified short phrases (as shown in Figure~\ref{fig:figure-main}) instead of the actual misrepresented study as evidence. 
In a real-world scenario the relevant parts of the misrepresented study are not presented in this form. Here, the models must first \emph{(i)}~identify the relevant passages within the entire misrepresented study and \emph{(ii)}~verbalize the fallacious reasoning based on the original scientific text. To address this, we present \dataset{} (§\ref{sec:dataset}), an extension of \missci{} with 2,257 human-annotated links between the simplified phrases from \missci{} and the real-world passages from the misrepresented publications.

Figure~\ref{fig:figure-main} shows (parts of) a fallacious logical argument from \missci{} for the claim that ``\emph{hydroxychloroquine is a cure for COVID-19}''.  Each claim in \missci{} has a kernel of truth which is anchored in the study's content (``\emph{Chloroquine reduced infection of the coronavirus}''), denoted as the \emph{accurate premise} in green. 
However, additional information from the same study  (``\emph{The study used cell cultures for their experiments}'') undermines the claim's validity and reveals a reasoning gap. The relevant content from the study that is needed to identify this reasoning gap between claim and study is referred to as the \emph{publication context}. The reasoning gap between the study's content and the claim as a conclusion indicates the fallacious reasoning. 
To bridge this reasoning gap 
the model must verbalize a \emph{fallacious premise} (``\emph{The results can be transferred to humans because the human body consists of cell structure.}''), blue in Figure~\ref{fig:figure-main}, and classify the fallacy associated with this fallacious premise (``\emph{fallacy of composition}'').  Unlike \missci{}, where the accurate premise and publication context are provided directly, \dataset{} requires LLMs to retrieve the required real-world passages from the misrepresented study and reason over them. A complete fallacious argument, including the misrepresented study passages, is provided in §\ref{appendix:example-argument}.
\dataset{} is the first fallacy dataset to pair real-world misinformation claims with publication passages as evidence. This setup is identical to the input used by evidence-based scientific AFC models and enables their evaluation over real-world fallacious misinformation. 
We use \dataset{} to answer the following research questions: 
\begin{enumerate}[noitemsep]
    \item How well can existing ranking approaches select the required publication passages (§\ref{sec:retrieval})?
    \item How well can LLMs reconstruct fallacious arguments using original studies compared to the simplified content in \missci{} (§\ref{sec:argument-reconstruction})? 
    \item Can evidence-based scientific AFC models use the content of misrepresented publications to detect the claims as misinformation (§\ref{sec:missci-plus-afc})?
\end{enumerate}
Our findings suggest that lexical and semantic similarity based ranking models perform the best at identifying the required evidence passages. However, all models cannot leverage these passages to refute  misinformation.
In summary, we contribute 
 \dataset{}, an extension of \missci{} to ground fallacies in real-world evidence, which bridges the gap to AFC. We propose novel task definitions, along with extensive experiments on retrieving the required passages %
 from the misrepresented study %
 needed
 to detect misrepresented publications in the wild,
as well as 
experiments using AFC models and LLMs in detecting science distortions.

\section{Related Work}
\label{sec:related-work}
\paragraph{Fallacy detection}
Much research on fallacy detection has primarily focused on surface-level fallacies \citep{habernal-etal-2017-argotario, habernal-etal-2018-name, da-san-martino-etal-2019-fine, sahai-etal-2021-breaking, piskorski-etal-2023-multilingual, salman2023detecting}. Other works extended these inventories to include logical fallacies that may require additional context for detection. 
However, all of this research relied on educational examples, fake news websites \citep{jin-etal-2022-logical}, or fact-checking articles \citep{musi2022developing, alhindi-etal-2022-multitask} and assumed that the explicitly stated text was sufficient to detect the fallacies. \missci{}~\citep{glockner2024dismantling} developed models to verbalize the implicit fallacious reasoning.
 \dataset{} differs from existing fallacy datasets by grounding implicit fallacies in real-world evidence documents.

\paragraph{Scientific AFC} 
A large body of research on scientific AFC used scientific documents as evidence to assess the veracity of claims \citep{wadden-etal-2020-fact, saakyan2021covidfact, sarrouti-etal-2021-evidence-based, kotonya-toni-2020-explainable, lu2023scitab,vladika-matthes-2023-scientific, vladika2024comparing}. These approaches face challenges with fine-grained differences, such as specificity mismatches \citep{wadden-etal-2022-scifact}. Our work bridges the gap between scientific AFC and fallacy detection and sheds light on the abilities of AFC models to reason over claims and misrepresented evidence passages.

\paragraph{Science communication}A very related research area concerns science communication research \citep{augenstein-2021-determining}, which compares claims and the cited evidence document across various dimensions such as claim strength \citep{li-etal-2017-nlp}, certainty of the used language \citep{pei-jurgens-2021-measuring}, sentence-level causal exaggerations \citep{yu-etal-2020-measuring}, quantification of the information match \citep{wright-etal-2022-modeling}, or combinations of multiple dimensions \citep{wuhrl2024understanding}.
Our work differs by focusing on harmful misinformation and tasks that involve retrieving relevant passages and articulating the fallacious reasoning.

\section{Background}
\subsection{Preliminaries}
\missci{}~\citep{glockner2024dismantling} comprises 208 inaccurate health-related claims that misrepresent scientific publications. 
Each claim is modeled as a fallacious logical argument where a claim is the (wrong) conclusion of its premises. Formally, each argument comprises exactly one accurate premise ($p_0$) based on which the claim was made, and at least one fallacious premise $\overline{p}_i$ with the fallacy class $f_i$ (cf. §\ref{appendix:fallacies} for details of all nine fallacy classes), that is needed to (falsely) conclude the claim $\overline{c}$ from the study's content. 
We refer to the pair of fallacious premise and the applied fallacy class ($\overline{p}_i$, $f_i$) as \emph{fallacy}. Each fallacy in \missci{} is linked to one publication context ($s_i$). The publication context contains the content of the misrepresented study, that necessitates the fallacy. 
For example, knowing that the study's observations were limited to cell cultures (publication context, orange, Figure~\ref{fig:figure-main}) reveals that the claim about the effectiveness (in humans) is unjustified. We define this disconnect between study content and its purported conclusions as a \emph{reasoning gap}, which fallacious premises attempt to bridge.
Each publication context (and accurate premise $p_0$) faithfully summarizes parts of the study.
\missci{} defines the argument reconstruction task as:
Given the publication context ($s_i$), the incorrect claim ($\overline{c}$) and the accurate premise ($p_0$), the model must verbalize the fallacious premise and detect the applied fallacy class ($\overline{p}_i$, $f_i$).

\definecolor{mylightgray}{rgb}{0.95,0.95,0.95} %
\definecolor{mydarkgray}{rgb}{0.6,0.6,0.6} %
\definecolor{mydarktext}{rgb}{0.25,0.25,0.25} 
\begin{figure}[h]
    \centering
    \setlength{\fboxsep}{10pt} %
    \tikz\node[draw=mydarkgray,fill=mylightgray,rounded corners=5pt,inner sep=10pt,text width=0.90\linewidth,align=justify] {
        \footnotesize
        \textcolor{mydarktext}{
In order to investigate the antiviral properties of chloroquine on SARS-CoV after the initiation of infection, Vero E6 cells were infected with the virus and fresh medium supplemented with various concentrations of chloroquine was added immediately after virus adsorption. Infected cells were incubated for an additional 16-18 h, after which the presence of virus antigens was analyzed by indirect immunofluorescence analysis. When chloroquine was added after the initiation of infection, there was a dramatic dose-dependant decrease in the number of virus antigen-positive cells (Fig. 2A). As little as 0.1-1 muM chloroquine reduced the infection by 50\% and up to 90-94\% inhibition was observed with 33-100 muM concentrations (Fig. 2B). At concentrations of chloroquine in excess of 1 muM, only a small number of individual cells were initially infected, and the spread of the infection to adjacent cells was all but eliminated. A half-maximal inhibitory effect was estimated to occur at 4.4 ± 1.0 muM chloroquine (Fig. 2C). These data clearly show that addition of chloroquine can effectively reduce the establishment of infection and spread of SARS-CoV if the drug is added immediately following virus adsorption.
        }
    };
    \caption{A real-world passage \citep{vincent2005chloroquine} communicates the paraphrased content $s_1$ from \missci{} that \emph{the study used cell cultures for their experiments}.}
    \label{fig:real-world-evidence}
\end{figure}

\subsection{Linked Passages}
\label{sec:background:missciplus}
 In \missci{}, the accurate premise ($p_0$) and publication contexts ($s_i$) were manually paraphrased from the HFC article and \emph{not} the study itself.
 Since these HFC articles were specifically written to explain to non-experts why the misrepresented study does not support the claim, the paraphrased publication contexts often reveal the reasoning gaps easily.
This severely limits the applicability of \missci{} in the real world, where models must identify relevant content from the entire misrepresented study, and  reason through complex scientific text.
To address this, \dataset{} links the simplified paraphrased information from \missci{} with actual passages from the misrepresented study. 
Figure~\ref{fig:real-world-evidence} shows a verbatim passage from the misrepresented study in \dataset{}, which is linked (i.e., communicates the same content) to the paraphrased content that ``\emph{the study used cell cultures for their experiments}'' from \missci{} (see Figure~\ref{fig:figure-main}).
Formally, given a misrepresented study $S$ with its passages $S_j \in S$, we linked the passage $S_j$ to the corresponding paraphrased information ($p_0$ and $s_i$) if (parts of) $S_j$ entail the paraphrased information ($p_0$ or $s_i$). Any passage $S_j$ linked to $p_0$ (or $s_i$) can replace $p_0$ (or $s_i$) during the argument reconstruction. The same passage may be linked to multiple paraphrased information and vice versa. We denote a passage $S_j$ as $S_j^0$ if it links to $p_0$.

\subsection{Subtasks}
We consider three sub-tasks for reconstructing fallacious arguments in the wild. First (§\ref{sec:subtask:1}), the model must retrieve a passage $S_j^0$, upon which the claim is based. This is crucial for understanding the general reasoning of the claim. 
Second (§\ref{sec:experiments:subtask2}), the model must retrieve all additional passages $S_j$ required to detect fallacies, i.e., passages linked to any publication context $s_i$. 
Lastly (§\ref{sec:argument-reconstruction}),  the argument reconstruction task is adapted from \missci{}, but replaces the paraphrased content with the respective linked passages.
In reality, each subtask relies on the output of preceding subtasks. In this work, we aim to establish a strong foundation for each sub-task individually and assume oracle input for each, laying the groundwork for a more robust end-to-end system in the future.

\section{Grounding \dataset{} with Evidence}
\label{sec:dataset}

To create
\dataset{}, we selected all fallacious logical arguments from \missci{}, for which the full misrepresented study is available via PMC\footnote{\url{https://www.ncbi.nlm.nih.gov/pmc/}}. This resulted in 118 fallacious arguments misrepresenting 100 distinct publications, which we automatically split into the constituent passages (cf. §\ref{appendix:dataset:step3:passage-selection:passage-extraction}).
We used the IMS model \citep{wright-etal-2022-modeling} to pre-select relevant passages and avoid exhaustively annotating every paraphrased information with every passage. The IMS model quantifies the information match between textual statements and scientific text, which aligns well with our needs.
For each paraphrased information ($p_0$ and $s_i$), we selected the top-ranked passage according to IMS  (cf. §\ref{appendix:dataset:step3:passage-selection:passage-selection}) and
  collected a minimum of six passages per argument in total, if possible. 
We employed two biology master's students with annotation experience in biomedical misinformation on \missci{}.
The annotators assigned an entailment label by comparing each paraphrased information ($p_0$ and $s_i$) with each selected passage ($S_j$),
determining whether $S_j$ entails (and hence is linked to) the paraphrased information.
Following \citet{glockner2023ambifc}, the annotators could express uncertainty if the entailment relation was ambiguous. If the paraphrased information was not linked to any pre-selected passage after consolidation, one annotator manually selected a corresponding passage from the entire study, if possible, which was then double-annotated.
We removed four arguments for which no paraphrased information could be linked to any passage, yielding 2,257 double-annotated relations between passages and paraphrased information across 114 arguments. 
We retained the same instances for validation (30 arguments) and test (84 arguments) splits as in \missci{}. 
The inter-annotator agreement, measured by Cohen's $\kappa$, was 0.602.
For details about the annotation process please refer to §\ref{appendix:dataset:step3:annotations}.
Overall, 400 pieces of paraphrased information (88.6\% of the accurate premises $p_0$; 72.0\% of the publication contexts $s_i$; 76.8\% overall) could be linked to at least one  passage (analysed in §\ref{sec:dataset:missing-links}). 
Descriptive statistics about the passages are in §\ref{appendix:passage-length}.

\section{Retrieving Relevant Passages}
\label{sec:retrieval}
A prerequisite to reconstruct the fallacious argument is the identification of the relevant passages in the study.
These passages provide evidence why the claim was made, and why this involves fallacies.
They are needed for the fallacious argument reconstruction and for effective debunking, which must explain why a claim was thought to be true and why it actually is not \citep{lewandowsky2020debunking}.

\subsection{Subtask 1: Finding the Kernel of Truth}
\label{sec:subtask:1}
Given an incorrect claim $\overline{c}$ that misrepresents a publication $S = [S_0, S_1, ..., S_n]$, the model must rank all passages $S_i \in S$ such that the top-ranked passage communicates the accurate premise ($p_0$) based on which the claim $\overline{c}$ was made (denoted as $S_j^0$). 
This task is similar to finding supporting evidence in automated fact-checking \citep{thorne-etal-2018-fever}, but differs as the evidence passage is in a ``corrupted'' support relationship with the claim, meaning it only supports the claim when a fallacy is involved. The passage $S_j^0$  explains the basis of the claim and reveals the (broken) rationale behind the claim.
We report P@$1$ and MRR 
over the subset of annotated passages with comprehensive annotations (\emph{closed}) and over all passages (\emph{open}).
The open evaluation is more realistic, but only serves as a lower bound. 

As \emph{baselines}, we randomly shuffled passages (\emph{random}) or maintained their original order from the publication (\emph{ordered}).
We used BM25 for lexical similarity-based ranking. 
As semantic \emph{embedding} based approaches, we ranked the passages by their cosine similarity to the claim using sentence embeddings from BioBERT~\citep{lee2020biobert}, fine-tuned by \citet{deka2022evidence} for evidence selection in scientific AFC, and prompt-based embeddings from INSTRUCTOR~\citep{su2022one}.
We also report the performance of the IMS \citep{wright-etal-2022-modeling}  used during the dataset construction.
Further, we trained DeBERTaV3~\citep{he2022debertav3} \emph{AFC} models on three scientific AFC datasets \textsc{SciFact}~\citep{wadden-etal-2020-fact}, \textsc{CovidFact}~\citep{saakyan2021covidfact}, \textsc{HealthVer}~\citep{sarrouti-etal-2021-evidence-based}, and their union, denoted as \emph{all} (cf. §\ref{appendix:scientific-afc}). Given the claim and the passages, we rank passages based on the predicted label probability for \textsc{Supported}, which fits closest to the task definition of finding a passage that seemingly supports the claim.

Finally, for LLM-based ranking we implemented PRP~\citep{qin2023large}, which reorders passages through pairwise comparisons akin to the early iterations of the bubble sort.  We used GPT-4~\citep{openai2023gpt4} and GPT-3.5 as proprietary LLMs, Llama3-8B~\citep{dubey2024llama} and Llama2-70B~\citep{touvron2023llama} as open-source LLMs. 
Implementation details are outlined in §\ref{appendix:subtask1:implementation}. 
Due to the high computational costs of PRP with growing numbers of documents, we only evaluate the  LLMs in the closed evaluation with prompt selection based on the development set, Table~\ref{tab:results:subtask1:llm-pairwise-prompt-tuning}). Preliminary experiments showed cheaper LLM-based methods like list-wise ranking were inefficient due to incorrect outputs and order sensitivity, which is a known limitation \citep{zhu2023large}.

\begin{table}[]
\small
    \centering
    \resizebox{\linewidth}{!}{%
    \begin{tabular}{ll | cc | c}
    \toprule
    && \multicolumn{2}{c|}{\emph{closed}} & \emph{open} \\ 
    &\textbf{Model} & \textbf{P@1} & \textbf{MRR}  & \textbf{MRR} \\
    \toprule
    &\emph{Random} & 0.360 & 0.566  & 0.209 \\
    \textbf{(1)} & \emph{Ordered} & 0.480 & 0.658  &  0.443\\
     & BM25 & 0.547 & 0.705 & 0.539\\
     \midrule

    & BioBERT ST & 0.547 & 0.712  &  0.582 \\

  \textbf{(2)}   &INSTRUCTOR  & 0.573 & 0.738  &  0.631 \\
    & IMS & 0.587 & 0.742   & \textbf{0.664}\\
    
     \midrule
    & AFC (SciFact) & 0.603 & 0.748  & 0.535\\
 \multirow{2}{*}{\textbf{(3)}}  &  AFC (CovidFact)  & 0.517 & 0.691 & 0.450 \\
   &  AFC (HealthVer) &  0.608 & 0.765  & 0.516 \\
    & AFC (\emph{all}) & 0.608 & 0.768  &  0.514\\

     \midrule
    & Llama2-70B & 0.711 & 0.830  & -- \\
\multirow{2}{*}{\textbf{(4)}}    & Llama3-8B& 0.729 & \textbf{0.850}  & -- \\
& GPT-3.5  & 0.671 & 0.815  & -- \\
&     GPT-4  & \textbf{0.742} & \textbf{0.850} & --\\
     \bottomrule
     
    \end{tabular}
    }
    \caption{Finding a passage $S_j^0$ based on which the claim was made \emph{(a)} among the annotated passages only (\emph{closed}) or \emph{(b)}  across the entire publication (\emph{open}). We list results for (\textbf{1}) baselines, \textbf{(2)} embedding rankers, \textbf{(3)} AFC models and  \textbf{(4)} LLMs. Averaged over five seeds (three for LLMs).}
    \label{tab:results:subtask1-use}
\end{table}

Table~\ref{tab:results:subtask1-use} shows solid performance across all models. The strong \emph{ordered} baseline suggests that claims often rely on early parts of a study. In the open evaluation, embedding-based approaches perform the best.
Note that the IMS model preselected the passages for annotation, 
and its performance must be interpreted with caution.
AFC models are superior to embedding models in the closed evaluation, but fall behind PRP ranking via LLMs.

\subsection{Subtask 2: Finding Undermining Passages}
\label{sec:experiments:subtask2}
Given an incorrect claim $\overline{c}$ that misrepresents a publication $S = [S_0, S_1, ..., S_n]$, and a passage $S_j^0 \in S$, based on which the claim was made, the model must rank all passages $S_i \in S$ such that the top-ranked passages $S_i$ expose reasoning gaps (i.e., they are linked to the publication context $s_i$) between the study content $S$ and the inaccurate claim $\overline{c}$.
This passage ranking task differs substantially from evidence retrieval in AFC, as the model must \emph{i}) understand the rationale behind the claim based on the accurate premise in $S^0_j$, and \emph{ii}) evaluate how each passage $S_i \in S$ to be ranked impacts this rationale.
For example, ``\emph{in vitro experiments}'' only indicate a fallacy because the claim relies on the results of these experiments and incorrectly transfers them to humans. 
This task also differs from multi-hop reasoning \citep{jiang-etal-2020-hover, ma-etal-2024-ex}, which connects information to \emph{establish} a reasoning path, e.g., to support or contradict a claim. Instead, the task identifies passages that \emph{disrupt} the reasoning behind the misinformation. %

We report the passage-level mean average precision (MAP) as our main metric.
We further report P@$1$ because one detectable fallacy is the minimum requirement to reject a claim. To measure how many distinct reasoning gaps can be detected, we report the fallacy-level recall of the top ten ranked passages (Fall-R@$10$). This penalizes models that only detect passages linked to the same publication context $s_i$ and, hence, can only detect the same subset of fallacies. We only evaluate this subtask on the \emph{open} subset, as most annotated passages are linked with reasoning gaps as per the dataset construction. 
Here, we slightly adapted the AFC-based ranking to use the sum of the predicted probabilities for the labels \textsc{Supported} and \textsc{Refuted}. We further assume oracle outputs from the previous subtask and prepended a randomly sampled gold $S_j^0$ to the claim in all baselines. These design choices follow our experiments on the validation split (cf. §\ref{appendix:subtask2}).

To solve this task, the model must first understand the (false) rationale behind the claim as expressed in the passage $S_j^0$ (e.g., communicating that ``\emph{chloroquine reduced infection of the coronavirus}''). Then, the model can identify how a different passage interacts with this reasoning to highlight a reasoning gap (e.g., that this was observed in ``\emph{in vitro}'' experiments). 
Table~\ref{tab:results:subtask2-open-use} shows 
that cosine-similarity-based ranking outperforms all AFC-based models that can jointly encode the evidence passage with the claim, despite the complexity of the reasoning required.
This suggests that lexical or semantic similarities correlate sufficiently strongly with passages that indicate reasoning gaps, while the acquired reasoning by AFC models %
seem to be not helpful. 

\begin{figure}[h]
\small
    \centering
    \includegraphics[width=\linewidth]{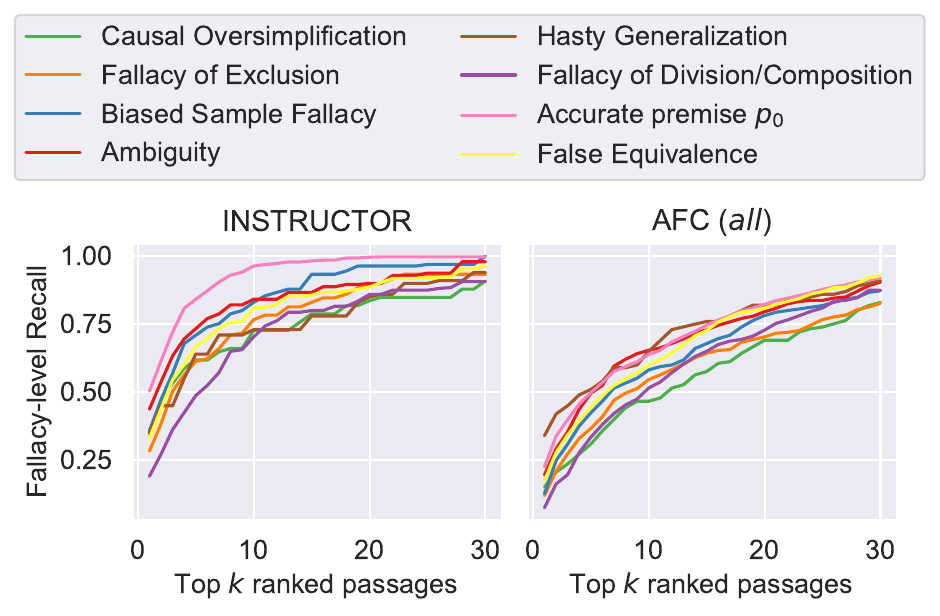}
    \caption{Recall of undermining passages per fallacy class (and accurate premise) over the top $k$ ranked passages. We only list fallacies with $\geq 20$ occurrences.}  
    \label{fig:figure-st2-afc-instructor-per-fallacy}
\end{figure}

\begin{table}[h]
\small
    \centering
    \begin{tabular}{ll|cc|c}
    \toprule
    &\textbf{Model} & \textbf{MAP} & \textbf{P@1} & \textbf{Fall-R@10} \\
    \toprule

    & \emph{Random}  & 0.205 & 0.136 & 0.438\\
    \textbf{(1)} & \emph{Ordered}  & 0.286 & 0.298 & 0.497\\
     & BM25  & 0.496 & 0.617  & 0.602\\
    \midrule
     &INSTRUCTOR   & \textbf{0.541} & \textbf{0.652}  & \textbf{0.613}\\
    \textbf{(2)} &SPICED-IMS & 0.524 & 0.640  & 0.595\\
    &BioBERT ST  & 0.491 & 0.600 & 0.570\\
    
    \midrule
    &AFC (SciFact) & 0.360 & 0.326 & 0.518\\
    \multirow{2}{*}{\textbf{(3)}}
    &AFC (CovidFact) & 0.380 & 0.457 & 0.538\\
    &AFC (HealthVer)  & 0.368 & 0.410 & 0.544\\
    &AFC (\emph{all})   & 0.306 & 0.338 & 0.554\\
    \bottomrule
    \end{tabular}
    \caption{Retrieving passages linked to fallacies from the entire publication (\emph{open}) for \textbf{(1)} baselines, \textbf{(2)} embedding rankers and \textbf{(3)} AFC models. Avg over five seeds.}
    \label{tab:results:subtask2-open-use}
\end{table}

We show the recall of detectable fallacy classes over the number of retrieved passages for the INSTRUCTOR and AFC (all) in Figure~\ref{fig:figure-st2-afc-instructor-per-fallacy}. Overall, within the same model, different fallacy classes follow similar trends. The superior performance of INSTRUCTOR seemingly relies on fallacies that can be detected from $S_j^0$ passages, which implies that INSTRUCTOR does not really capture the problematic nature of the passages that undermine the claim. For comparison, we visualize the performance without $S_j^0$ passages in §\ref{appendix:subtask2:fallacy-analysis}, Figure~\ref{fig:figure-st2-afc-instructor-per-fallacy-nop0}, which reduces the performance gap between the two ranking models.

\section{Subtask 3: Fallacious Argument Reconstruction}
\label{sec:argument-reconstruction}
Given an incorrect claim, $\overline{c}$, and relevant selected passages $S_j \in S$ from the misrepresented study $S$ that contain the accurate premise $p_0$ and the necessary publication contexts $s_i$ for detecting fallacies,  the model must generate the fallacious premises $\overline{p}_i$ along with their corresponding fallacy classes $f_i$. These generated fallacious premises, $\hat{\overline{P}} = [\hat{\overline{p}}_1, \hat{\overline{p}}_2, ..., \hat{\overline{p}}_n]$, should bridge all reasoning gaps between $S$ and the incorrect claim $\overline{c}$. 
We assume oracle output of both passage retrieval tasks from Section~\ref{sec:retrieval} and only provide passages linked to a fallacy (with at least one $S_j^0$ passage as the basis for the claim). We experimented with GPT-4 Turbo and GPT-3.5 as  proprietary LLMs , and Llama3-8B as a small open-source LLM with strong leaderboard performance.
As in \missci{}, we prompted the LLMs in a zero-shot setting (cf. §\ref{appendix:llm-argument-reconstruction-details} for prompt selection).
By default, all prompts include \textbf{D}efinition, \textbf{L}ogical Form and toy \textbf{E}xamples, as shown in Figure~\ref{fig:dle-example}, from literature \citep{bennett2012logically, cook2018deconstructing} as supplementary  information for each fallacy  (cf. §\ref{appendix:fallacies} for all fallacy classes).

\begin{figure}[]
    \centering
    \small
    \begin{tabularx}{\linewidth}{X}
                \toprule
                Fallacy class: \emph{Fallacy of Composition} \\
                \midrule
            \textbf{Definition}:
            Inferring that something is true of the whole from the fact that it is true of some part of the whole \\
            \midrule
            \textbf{Logical Form}:
            A is part of B. A has property X. Therefore, B has property X. \\
            \midrule
            \textbf{Example}:
            Hydrogen is not wet.  Oxygen is not wet.  Therefore, water (H2O) is not wet. \\
            
            \bottomrule
        \end{tabularx}
    \caption{Examples for the (\textbf{D})efinition, (\textbf{L})ogical form and (\textbf{E})xample for the \emph{Fallacy of Composition}, used as supplementary fallacy information in the prompts.}
    \label{fig:dle-example}
\end{figure}

\subsection{Holistic Argument Evaluation}

The evaluation on \dataset{} faces two challenges:
First, unlike \missci{}, there is no one-to-one mapping between fallacies and passages. This is because a single passage can simultaneously contain multiple publication contexts ($s_i$), each linked to different fallacies, and a single fallacy can also relate to multiple passages.
Second, different fallacies ($\overline{p}_i$, $f_i$) that share the same fallacy class ($f_i$) but address different reasoning gaps between the study and the claim may be present.
For example, the claim that ``\emph{hydroxychloroquine is a cure for COVID-19}'' in Figure~\ref{fig:figure-main} is based on two false assumptions: \emph{i}) that hydroxychloroquine will have the \emph{same effect as chloroquine}, and \emph{ii}) that SARS-CoV-2 will \emph{behave in the same way as SARS-CoV-1}.  Each assumption presents a separate error in the claim, both of which are based on the \emph{False Equivalence} fallacy.
Hence, during evaluation, the fallacy class alone is insufficient to determine which of the two problems the model addressed with its generated fallacies. To match the generated fallacies with the gold fallacies, it is essential to additionally use the fallacious premises.

To address these challenges, we evaluated the fallacious argument holistically. 
Given \emph{all} relevant passages of the misrepresented study, we expect up to five ranked fallacies ($\hat{\overline{p}}_i$, $\hat{f}_i$), where five is the maximum number of distinct reasoning gaps, that must be addressed with fallacies, in \missci{}.
Following \newcite{schlichtkrull2023averitec}, we automatically match the generated fallacies with the gold fallacies at the \emph{argument} level (instead of evaluating fallacies per publication context $s_i$ as in \missci{}). 
We define a function $\phi: (y, \hat{y}) \rightarrow \{0,1\}$ which discerns if the predicted  fallacy $\hat{y}_i$ = ($\hat{\overline{p}}_i$, $\hat{f}_i$) and the gold fallacy $y_k$ = ($\overline{p}_k$, $f_k$) match based on two implementations:
\begin{itemize}[noitemsep]
    \item $\phi^{\mathrm{f}}$ outputs $1$ if the predicted fallacy class equals the gold fallacy class.
    \item $\phi^{\mathrm{f+p}}$ outputs $1$ if the generated fallacious premise additionally bridges the same reasoning gap as the gold fallacious premise.
\end{itemize}
The $\phi^{\mathrm{f}}$ is an upper bound, as it does not penalize models for poorly phrased premises. $\phi^{\mathrm{f+p}}$ uses a Llama3-8B model, fine-tuned with QLoRA adapters \citep{dettmers2023qlora} on the human evaluation data from \missci{} (cf. §\ref{subtask3-judge}). The accuracy via cross-validation is 79.8\% (78.8 in F1-macro).
We do not perform a human evaluation, which is too complex given the many-to-many relationship between the predicted and the gold fallacies, and is hard to reproduce for future work. Instead, we report evaluation measures based on the two complementary implementations $\phi$, which we deem adequate to answer our research questions.
As a primary measure, we report the recall of reasoning gaps for which $\phi$ found a match among the five fallacy predictions (R@$5$).
Following \citet{glockner2024dismantling}, we use precision P@$1$, to check if the top-ranked fallacy is correct, and Arg@$1$, which considers an argument as successfully rejected if at least one of the predicted fallacies is correct.

\begin{table}[t]
\small
    \centering
    \resizebox{\linewidth}{!}{%
    \begin{tabular}{ll|ccc}
    \toprule
    \textbf{LLM} & \textbf{Passages} & \textbf{R@5} & \textbf{P@1} & \textbf{Arg@1}  \\
    \midrule

    Llama3-8B & per-passage & \textbf{0.226} & \textbf{0.290} & \textbf{0.476} \\
    Llama3-8B & all passages & 0.199 & 0.266 & 0.425 \\
    GPT-3.5  & per-passage & 0.165 & 0.190 & 0.361 \\
    GPT-3.5 & all passages & 0.089 & 0.194 & 0.206 \\

    \bottomrule

    \end{tabular}
    }
    \caption{\emph{All passage}  prompting vs. \emph{per passage} prompting. Averaged over three seeds.}
    \label{tab:subtask3-argument-reconstruction}
\end{table}

\begin{table*}[h]
\small
    \centering
    \begin{tabular}{ll|ccc|ccc} \\
    \toprule
      && \multicolumn{3}{c|}{\missci{}} & \multicolumn{3}{c}{\dataset{}} \\
      \textbf{LLM} & \textbf{Info} & \textbf{R@5} ($\phi^{\mathrm{f+p}}$) & \textbf{R@5} ($\phi^{\mathrm{f}}$) &\textbf{Arg@$1$ ($\phi^{\mathrm{f+p}}$)} & \textbf{R@5} ($\phi^{\mathrm{f+p}}$) & \textbf{R@5} ($\phi^{\mathrm{f}}$) & \textbf{Arg@$1$ ($\phi^{\mathrm{f+p}}$)}\\
      \midrule
      & DLE & \textbf{0.277} & \textbf{0.514} & \textbf{0.552} & \textbf{0.226} & \textbf{0.477} & \textbf{0.476} \\
      \multirow{ 2}{*}{Llama3-8B} & DL & 0.241 & 0.445 & 0.512 & 0.195 & 0.463 & 0.413 \\
       & DE & 0.227 & 0.470 & 0.480 & 0.174 & 0.449 & 0.389 \\
        & LE & 0.255 & 0.469 & 0.504 & 0.209 & 0.439 & 0.460 \\
        \midrule
    & DLE & 0.248 & 0.491 & 0.512 & \textbf{0.165} & \textbf{0.428} & \textbf{0.361} \\
      \multirow{ 2}{*}{GPT-3.5} & DL & 0.232 & 0.492 & 0.464 & 0.146 & 0.416 & 0.321 \\
       & DE & \textbf{0.276} & \textbf{0.517} & \textbf{0.567} & 0.160 & 0.400 & 0.333 \\
        & LE & 0.249 & 0.478 & 0.524 & 0.157 & 0.410 & 0.341 \\
                \midrule
    & DLE & \textbf{\underline{0.332}} & 0.486 & \textbf{\underline{0.619}} & 0.224 & 0.458 & 0.452 \\
      \multirow{ 2}{*}{GPT-4 Turbo} & DL & 0.308 & 0.500 & 0.583 & 0.238 & 0.495 & 0.488 \\
       & DE & 0.318 & \textbf{\underline{0.528}} & 0.595 & 0.210 & 0.491 & 0.440 \\
        & LE & 0.304 & 0.505 & 0.583 & \textbf{\underline{0.252}} & \textbf{\underline{0.519}} & \textbf{\underline{0.500}} \\

     \bottomrule
    \end{tabular}
    
    \caption{Argument reconstruction performance using paraphrased information from \missci{} compared to the real passages from \dataset{} across various fallacy information. Results (expect GPT-4) are averaged over 3 seeds. We evaluate different combinations of fallacy (\textbf{D})efinition, (\textbf{L})ogical form and (\textbf{E})xample in the prompt.}
    \label{tab:subtask3-missci-vs-us}
\end{table*}

\subsection{Experiments}
\label{sec:experiments:per-passage-prompting}
We compare the performance when prompting LLMs based on each passage $S_j$ individually (\emph{per-passage}) and when including all concatenated passages in a single prompt (\emph{all passages}). The \emph{per-passage} prompts follow  \missci{}, but replace the accurate premise ($p_0$) and publication context ($s_i$) with the respective real-world passages that contain the same information. %
We always provided oracle passages and selected the top five ranked fallacy predictions per argument for evaluation using $\phi^{\mathrm{f+p}}$ in Table~\ref{tab:subtask3-argument-reconstruction}.
Per-passage prompting shows superior performances for both LLMs, but is more expensive because it requires multiple prompts per argument. The higher (or similar) precision suggests that focusing on one passage (per-passage prompting) can be advantageous when identifying the fallacies. Llama3-8B outperforms GPT-3.5 across all measures,
but detects none of the annotated fallacies for more than half of the misinformation, according to Arg@$1$.

\label{sec:st3:comparison}
Table~\ref{tab:subtask3-missci-vs-us} compares the performance of all three LLMs and ablations over the different fallacy information in the prompts (\textbf{D}efinition, \textbf{L}ogical Form and \textbf{E}xplanation) on \missci{} and \dataset{} using per-passage  prompting (using the paraphrased $s_i$ and $p_0$ on \missci{}). GPT-4 performs best across all measures and datasets on this task. We further observe a considerable impact of different fallacy information for each LLM. Yet, none is universally beneficial or harmful for all tested LLMs.

A clear trend of decreasing performance from the paraphrased information in \missci{} to the real-world passages in \dataset{} is evident. We note that, the approximate measures via $\phi^{\mathrm{f}}$ and $\phi^{\mathrm{f+p}}$ may underestimate performance as they cannot match valid fallacies not covered by the annotations \citep{glockner2024dismantling}.
The performance from \missci{} to \dataset{} only drops marginally for Llama3-8B and GPT-4 Turbo. Interestingly,  GPT-3.5 outperforms Llama3-8B using the paraphrased content. However, its poor performance across all evaluated prompts on \dataset{} corroborates the findings in Table~\ref{tab:subtask3-argument-reconstruction} and suggests poor adaptability toward realistic scientific text on this task.

\section{Scientific AFC Evaluation}
\label{sec:missci-plus-afc}
A key novelty of \dataset{} are the claims paired with real-world evidence passages that can be assessed by fallacy detection models \emph{and} evidence-based fact-checking models. This allows to test AFC models on fallacious claims.

\subsection{Fine-Tuned Models for Scientific AFC}

\begin{table}[h]
\small
    \centering
    \resizebox{\linewidth}{!}{%
    \begin{tabular}{l| c c c c |  c}
    \toprule
    & \multicolumn{4}{c|}{\emph{Argument level}} & \emph{AFC} \\
   \textbf{Training Data} & \textbf{Sup.} & \textbf{Ref.} & \textbf{Mix.} & \textbf{NEI} & \textbf{Acc.}\\
   \toprule
    SciFact & 55.5 & 4.5 & 23.3 & 16.7&  88.9\\
    HealthVer& 39.0 & \textbf{20.0} & 27.1 & 13.8& 82.1\\
    CovidFact & \textbf{36.9} & 12.1 & \textbf{51.0} & --&  90.7\\
    \midrule
    \emph{All AFC} & 48.8 & 11.9 & 27.1 & 12.1&  84.1\\
    
    \bottomrule

    \end{tabular}
    }
    \caption{Veracity predictions from scientific AFC models on 84 misinformation claims with 510 evidence passages in \dataset{}. Averaged over five seeds.}
    \label{tab:afc-on-missciplus}
\end{table}

Given a claim $\overline{c}$ that misrepresents the publication $S$, we form $n$ fact-checking instances ($\overline{c}$, $S_j$) by treating each of the $n$ annotated passages $S_j \in S$, as evidence. We use the scientific AFC models %
from §\ref{sec:retrieval} to predict $n$ veracity labels.
Following \citet{schlichtkrull2023averitec}, we assign an overall veracity label for a claim as \textsc{Mixed} if the labels  \textsc{Supported} and \textsc{Refuted}  were predicted. If the model only predicted \textsc{NotEnoughInformation} (\textsc{NEI}), the overall verdict was \textsc{NEI}. We label all other cases in which the model predicted \textsc{Supported} (with optional \textsc{NEI}) or \textsc{Refuted} (with optional \textsc{NEI}) as \textsc{Supported} or \textsc{Refuted}, respectively.
The studies in \dataset{} constitute trustworthy evidence that do not support the inaccurate claims -- HFC have rated them to misrepresenting these studies. A literate scientific AFC model should equally detect the claims as misinformation. %

Table~\ref{tab:afc-on-missciplus} shows that all AFC models yielded high in-domain accuracy (82-90\%) on the respective AFC dataset. 
In \dataset{}, the misrepresented studies were (mis)used as evidence to back up inaccurate claims. Therefore, if the AFC model assigns any label other than \textsc{Supported}, it can be considered a correct rejection of the claim.
 Yet, AFC systems falsely predict 37-56\% of the claims to be true. 
 The seemingly best \textsc{CovidFact} model uses binary classification without \textsc{NEI} and benefits from a substantially increased chance to predict \textsc{Mixed}.

\begin{figure}[t]
\small
    \centering
    \includegraphics[width=1\linewidth]{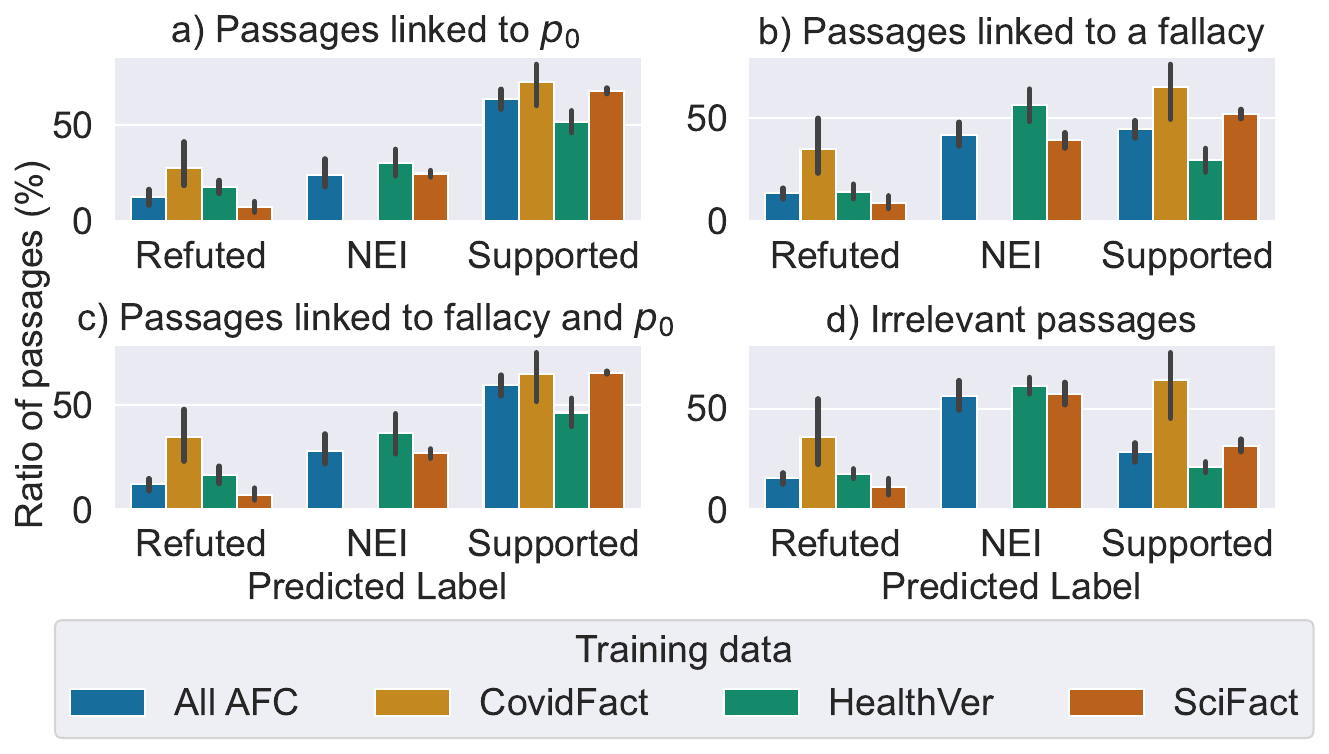}
    \caption{AFC predictions over passages linked to the accurate premise  (\emph{top left}), to reasoning gaps (\emph{top right}), to both (\emph{bottom left}) or none (\emph{bottom right}).
    }
    \label{fig:figure-afc-passage-distributions}
\end{figure}

Grounding the verdict of AFC models in the used evidence is critical for their  trustworthiness and often is a key part of the evaluation protocol \citep{thorne-etal-2018-fever, wadden-etal-2020-fact, glockner2023ambifc}. 
Table~\ref{tab:afc-on-missciplus} only shows if the LLM rejected a claim based on \emph{any} passage, not if the LLM assigned the correct label for the provided passage.
To understand whether the AFC models reject the claims based on the undermining evidence that indicates reasoning gaps, we visualize the AFC prediction over different passages in Figure~\ref{fig:figure-afc-passage-distributions}. Intuitively, the models mostly predict \textsc{Supported} over passages based on which the claim was made (\emph{top left}). If the same passage is additionally linked to a fallacy (\emph{bottom left}), the predicted distribution does not change much, suggesting unawareness of the fallacy.
Based on passages that are linked to a fallacy but not to the accurate premise (\emph{top right}), the distribution changes towards \textsc{NEI}. This is similar to completely unrelated passages (\emph{bottom right}), again suggesting unawareness of the scientific distortion.
More analysis is in §\ref{appendix:afc-on-missci}.

\subsection{LLMs as Scientific AFC Models}
HFCs disseminates their fact-checking articles widely, which can give unfair advantages to models when they are evaluated on the same misinformation \citep{glockner-etal-2022-missing}.
We prompted each LLM (\emph{a})~without any evidence to test whether it already has parametric knowledge about the claims from pretraining \citep{magar-schwartz-2022-data} and (\emph{b})~with the relevant evidence passages to test their reasoning capabilities in a \emph{RAG} style, similarly to \citet{manakul-etal-2023-selfcheckgpt}.
For parametric knowledge, we let the LLM generate a fact-checking article (\emph{FC}), which is closest to our application, but may bias the model to refute a claim due to the prevalence of misinformation in fact-checking articles. As a neutral prompt, we additionally \emph{asked} the LLM to predict the veracity of the claim to the best of their knowledge.
All prompts are listed in §\ref{appendix:prompts}.
Table~\ref{tab:llm-as-afc} shows the results on \dataset{} and 100 randomly sampled correct claims from \textsc{HealthVer} and \textsc{CovidFact} (50 per dataset). Both datasets contain real-world health-related claims from the internet.
The percentages may not add up to 100\% if LLMs refuse to answer.

\begin{table}[h]
\centering
\resizebox{\linewidth}{!}{%
\begin{tabular}{ll|ccc|ccc}
\toprule
& & \multicolumn{6}{c}{\emph{Predicted as}} \\
& \textbf{LLM} & \textbf{True} & \textbf{False} & \textbf{NEI} & \textbf{True} & \textbf{False} & \textbf{NEI}\\
\midrule
& Llama2 & 1.6 & 61.1 & 37.3 & 34.7 & 22.3 & 41.3 \\
Know & Llama3 & 0.0 & 86.9 & 2.4 & 20.0 & 43.3 & 14.3 \\
(\emph{FC}) & GPT4 & 0.0 & 85.3 & 14.7 & 59.0 & 23.0 & 17.3 \\
& GPT3.5 & 0.8 & 71.0 & 28.2 & 46.7 & 17.3 & 35.7 \\
\midrule
& Llama2 & 0.0 & 100.0 & 0.0 & 29.7 & 69.3 & 1.0 \\
Know & Llama3 & 8.3 & 88.9 & 2.4 & 68.7 & 26.0 & 3.0 \\
(\emph{Ask}) & GPT4 & 3.6 & 68.3 & 27.8 & 49.7 & 6.7 & 36.3 \\
& GPT3.5 & 1.6 & 50.4 & 48.0 & 47.3 & 6.0 & 45.0 \\
\midrule
& Llama2 & 23.8 & 61.5 & 12.7 & 58.7 & 29.7 & 10.7 \\
\multirow{2}{*}{RAG} & Llama3 & 44.4 & 53.2 & 2.4 & 80.3 & 16.3 & 3.3 \\
& GPT4 & 27.4 & 34.1 & 38.5 & 55.0 & 4.0 & 41.0 \\
& GPT3.5 & 38.9 & 31.7 & 29.0 & 78.0 & 5.3 & 16.0 \\
\midrule
\multicolumn{2}{c}{} & \multicolumn{3}{c|}{\textbf{Misinformation}} & \multicolumn{3}{c}{\textbf{True claims}} \\
\bottomrule
\end{tabular}
} %
\caption{Averaged veracity predictions from LLMs on misinformation from \dataset{} (\emph{left}) and true claims from \textsc{HealthVer} and \textsc{CovidFact} (\emph{right}).}
\label{tab:llm-as-afc}
\end{table}

All LLMs tend to have parametric knowledge about the veracity of many claims. 
When providing evidence, the ratio of claims predicted as \textsc{True} increases not only for correct claims but also for misinformation,
despite that the \emph{misrepresented} evidence does not truly support the claims. This phenomenon is similar to overruling the internal knowledge with evidence \citep{wu2024faithful} and suggests that the LLMs do not identify the reasoning gaps between the claim and the evidence. Similarly to humans, LLMs seem prone to misinterpreting scientific publications. Due to their high persuasiveness \citep{augenstein2023factuality, el2024mechanism}, this can lead to disastrous problems when people rely on LLMs, even if LLMs transparently output the (misrepresented) used evidence.

Akin to previous work on fallacy recognition \citep{jin-etal-2022-logical, alhindi-etal-2022-multitask, glockner2024dismantling}, the prompts in this work focus on \emph{which} fallacies apply instead of \emph{whether} any applies, and are no panacea either: 
in preliminary experiments (cf. §\ref{appendix:afc-binary-fallacy-detection}), LLMs found fallacies in 85-99\% of the claims in \dataset{} but also in 78-99\% of the true claims.
Overall, \dataset{} unifies fallacy detection with AFC and provides a resource for studying how to handle fallacious arguments and true claims alike.

\section{Conclusion and Future Work}

We introduced \dataset{}, an extension of \missci{} to reconstruct fallacious arguments based on real passages of the misrepresented studies. We showed that existing ranking models and LLMs struggle to reconstruct fallacious arguments using the real-world evidence. Moreover, fine-tuned 
AFC
models and LLMs failed to %
refute claims in \dataset{} when presented with misrepresented evidence, highlighting the dangers of persuasive LLMs. 
Future work may improve models using synthetic data generation approaches, or extend the task definition over multiple documents.

\newpage

\section*{Limitations}
\dataset{} assumes that the claims are based on a single publication only and that each publication is inherently trustworthy, i.e., that the only error was done in the reasoning from the publication to the claim. In the real world, finding complementary credible evidence is critical. We observed that sometimes single passages are insufficient to ground each fallacy in evidence. 
Currently \dataset{} does not address grounding fallacies in multi-modal content, cases that require evidence from multiple passages, or conclusions that can only be drawn from the analysis of the entire study.
We leave these challenges for future research. 
Our evaluation is based on automatic matching, which is inevitably imperfect. Models may detect fallacies that are valid but not contained in the original annotations in \missci{}. We compensate for this with two complementary matching strategies and experiments over different seeds and prompt variations to confirm the robustness of our observations. By focusing on recall, we further do not penalize models for %
predicting fallacies that are not within our annotations while still requiring the models to detect the most prominent ones as highlighted by the annotators based on the HFC articles in \missci{}.
Our results are reported over two representative state-of-the-art LLMs at the time of writing (Llama3 8B as an easy-to-run open-source LLM and GPT4 as a proprietary LLM), and our claims are bound to these models. We opted for extensive tuning of these models to establish strong baselines for the novel tasks rather than providing comprehensive comparisons across various LLMs. 
While our focus lies in grounding the individual argument constituents in the real-world misrepresented study, which contains 2,257 annotated links and 400 argument constituents, \dataset{} is only based on 114 fallacious arguments.
This may lead to variance and biases in the experiments, which must be interpreted with caution. We note that %
creating
high-quality fallacy datasets with complex fallacious arguments requires suitable, professionally fact-checked claims, for which data is scarce. Future work could explore synthetic data generation to help bridge this gap.

\section*{Ethics Statement}
The research questions targeted in this work aim to improve the detection of claims that distort scientific publications, which is ethically uncritical. Ethical concerns are bound to cases in which the content of this study are used in unintended ways.

\paragraph{Dual Use}
False interpretations of health-related claims can have disastrous effect. Any output of models derived from \dataset{} only serves research purposes to detect such misinformation, but under no circumstances must be considered accurate without consulting experts in the field. 
Our work poses dangers for dual use, particularly verbalizing the fallacious reasoning to draw incorrect conclusions from real-world studies. While generating (parts of) misinformation always poses a risk, it is unavoidable to build resilience against real-world misinformation, as demonstrated in previous work \citep{zellers2019defending, huang-etal-2023-faking, alhindi2023large, glockner2024dismantling}. 

\paragraph{Data Collection}
All publications used in \dataset{} have been published by the respective authors and we did not anonymize their work. 
All publications used in \dataset{} are openly available and are part of the public discourse; in fact, they have even been distorted by misinformation. Hence, similarly to other scientific corpora \citep{lehman-etal-2019-inferring, lo-etal-2020-s2orc} that rely on such publications as evidence, we did not ask for explicit permission from the authors of each study to use their work.

\section*{Acknowledgments}
 This research work has been funded by the German Federal Ministry of Education and Research and the Hessian Ministry of Higher Education, Research, Science and the Arts within their joint support of the National Research Center for Applied Cybersecurity ATHENE.
Yufang Hou was supported by the Visiting Female Professor Programme from TU Darmstadt.
We gratefully acknowledge the support of Microsoft with a grant for access to OpenAI GPT models via the Azure cloud (Accelerate Foundation Model Academic Research).
We are grateful to our dedicated annotators who helped to create \dataset{} and to the anonymous reviewers and meta reviewer for their valuable feedback.
Finally, we wish to thank Jan Buchmann, Sukannya Purkayastha, Luke Bates and Jonathan Tonglet for their valuable feedback on an early draft of this work. 

\newpage
\bibliography{cleanbib}
\appendix

\section{Fallacies in \dataset{}}
\label{appendix:fallacies}
\begin{table*}[]
\footnotesize{
    \centering
    \begin{tabularx}{\textwidth}{XX}
    \toprule
    \textbf{Definition} & \textbf{Logical Form} \\
    \toprule
    \multicolumn{2}{l}{\textbf{\textsc{Ambiguity}}} \\
    When an unclear phrase with multiple definitions is used within the argument; therefore, does not support the conclusion. & \textit{Claim X is made. Y is concluded based on an ambiguous understanding of X.}\\
    \midrule
    
    \multicolumn{2}{l}{\textbf{\textsc{Equivocation}} (merged with \textbf{\textsc{Ambiguity}})} \\
    When the same word (here used also for phrase) is used with two different meanings. Equivocation is a subset of the ambiguity fallacy. & \textit{Term X is used to mean Y in the premise. Term X is used to mean Z in the conclusion.} \\
    \midrule

    \multicolumn{2}{l}{\textbf{\textsc{Impossible Expectations}} / \textbf{\textsc{Nirvana Fallacy}}} \\
    Comparing a realistic solution with an idealized one, and discounting or even dismissing the realistic solution as a result of comparing to a “perfect world” or impossible standard, ignoring the fact that improvements are often good enough reason. & \textit{X is what we have. Y is the perfect situation. Therefore, X is not good enough.} \\
    \midrule

    \multicolumn{2}{l}{\textbf{\textsc{False Equivalence}}} \\
    Assumes that two subjects that share a single trait are equivalent. & \textit{X and Y both share characteristic A. Therefore, X and Y are [behave] equal. }\\
    \midrule

    \multicolumn{2}{l}{\textbf{\textsc{False Dilemma}}} \\
    Presents only two alternatives, while there may be another alternative, another way of framing the situation, or both options may be simultaneously viable. & \textit{Either X or Y is true.}  \\
    \midrule
    
    \multicolumn{2}{l}{\textbf{\textsc{Biased Sample Fallacy}}} \\
Drawing a conclusion about a population based on a sample that is biased, or chosen in order to make it appear the population on average is different than it actually is. & \textit{Sample S, which is biased, is taken from population P. Conclusion C is drawn about population P based on S.} \\
\midrule
    
    \multicolumn{2}{l}{\textbf{\textsc{Hasty Generalization}}}\\
    Drawing a conclusion based on a small sample size, rather than looking at statistics that are much more in line with the typical or average situation. &
    \textit{Sample S is taken from population P. Sample S is a very small part of population P. Conclusion C is drawn from sample S and applied to population P.}   \\
    \midrule

    \multicolumn{2}{l}{\textbf{\textsc{False Cause Fallacy}} (use as \textbf{\textsc{Causal Simplification})}} \\    
    Post hoc ergo propter hoc — after this therefore because of this. Automatically attributes causality to a sequence or conjunction of events. & \textit{A is regularly associated with B; therefore, A causes B.}  \\
    \midrule

\multicolumn{2}{l}{\textbf{\textsc{Single Cause Fallacy}} (use as \textbf{\textsc{Causal Simplification)}}} \\
Assumes there is a single, simple cause of an outcome. & \textit{X is a contributing factor to Y. X and Y are present. Therefore, to remove Y, remove X.} \\
\midrule

    \multicolumn{2}{l}{\textbf{\textsc{Fallacy of Composition}}} \\
    Inferring that something is true of the whole from the fact that it is true of some part of the whole. & \textit{A is part of B. A has property X. Therefore, B has property X.} \\
    \midrule

    \multicolumn{2}{l}{\textbf{\textsc{Fallacy of Division}} (merged with \textbf{\textsc{Fallacy of Composition})}} \\
    Inferring that something is true of one or more of the parts from the fact that it is true of the whole. & \textit{A is part of B. B has property X. Therefore, A has property X.} \\
    \midrule
    
    \multicolumn{2}{l}{\textbf{\textsc{Fallacy of Exclusion}} / \textbf{\textsc{Cherry Picking}} / \textbf{\textsc{Slothful Induction}}} \\
When only select evidence is presented in order to persuade the audience to accept a position, and evidence that would go against the position is withheld (Cherry Picking). Ignores relevant and significant evidence when inferring to a conclusion (Slothful Induction -- focus on neglect). & \textit{Evidence A and evidence B is available. Evidence A supports the claim of person 1. Evidence B supports the counterclaim of person 2. Therefore, person 1 presents only evidence A.}  \\

     \bottomrule
     
    \end{tabularx}
    }
    \caption{Fallacy Overview. Definition and logical form taken from \citet{bennett2012logically} and \citet{cook2018deconstructing}. Table as provided by \citet{glockner2024dismantling}.}
    \label{tab:fallacy-overview}
\end{table*} 

\begin{table*}[]
    \centering
    \footnotesize{
    \begin{tabularx}{\textwidth}{X}
    \toprule
    \textbf{\textsc{Ambiguity}} \\
    \textit{It is said that we have a good understanding of our universe.  Therefore, we know exactly how it began and exactly when.}
    \\
    \midrule
    
    \textbf{\textsc{Equivocation}}\\
\textit{A feather is light. What is light cannot be dark. Therefore, a feather cannot be dark.} \\
    \midrule

    \textbf{\textsc{Impossible Expectations}} / \textbf{\textsc{Nirvana Fallacy}} \\
\textit{Seat belts are a bad idea. People are still going to die in car crashes.} \\
    \midrule

    \textbf{\textsc{False Equivalence}}\\
 \textit{They are both Felidae, mammals in the order Carnivora, therefore there's little difference between having a pet cat and a pet jaguar.}\\
    \midrule

    \textbf{\textsc{False Dilemma}} \\
    \textit{I thought you were a good person, but you weren’t at church today.}  \\
    \midrule
    
    \textbf{\textsc{Biased Sample Fallacy}} \\
    \textit{Based on a survey of 1000 American homeowners, 99\% of those surveyed have two or more automobiles worth on average \$100,000 each.  Therefore, Americans are very wealthy.} \\
\midrule
    
    \textbf{\textsc{Hasty Generalization}}\\
    \textit{My father smoked four packs of cigarettes a day since age fourteen and lived until age sixty-nine.  Therefore, smoking really can’t be that bad for you.}   \\
    \midrule

    \textbf{\textsc{False Cause Fallacy}}  \\    
\textit{Every time I go to sleep, the sun goes down.  Therefore, my going to sleep causes the sun to set. }  \\
    \midrule

\textbf{\textsc{Single Cause Fallacy}}\\
\textit{Smoking has been empirically proven to cause lung cancer. Therefore, if we eradicate smoking, we will eradicate lung cancer. } \\
\midrule

    \textbf{\textsc{Fallacy of Composition}} \\
\textit{Hydrogen is not wet.  Oxygen is not wet.  Therefore, water (H2O) is not wet.} \\
    \midrule

    \textbf{\textsc{Fallacy of Division}}  \\
\textit{His house is about half the size of most houses in the neighborhood. Therefore, his doors must all be about 3 1/2 feet high.} \\
    \midrule
    \textbf{\textsc{Fallacy of Exclusion}} / 
    \textbf{\textsc{Cherry Picking}} /  \textbf{\textsc{Slothful Induction}} \\
 \textit{Employer: ``It says here on your resume that you are a hard worker, you pay attention to detail, and you don’t mind working long hours.''}\\ \textit{Andy: ``Yes sir.''}\\ \textit{Employer: ``I spoke to your previous employer.  He says that you constantly change things that should not be changed, you could care less about other people’s privacy, and you had the lowest score in customer relations.''}\\
 \textit{Andy: ``Yes, that is all true, as well.''}  \\

     \bottomrule
     
    \end{tabularx}}
    \caption{Fallacy Examples (taken from \citet{bennett2012logically}). Table as provided by \citet{glockner2024dismantling}.}
    \label{tab:fallacy-overview-examples}
\end{table*} 

We provide the definitions, logical form, and examples from literature taken from  \citet{glockner2024dismantling} in Tables~\ref{tab:fallacy-overview}-\ref{tab:fallacy-overview-examples}.
\missci{} aggregates the fallacies into nine different classes:
\begin{enumerate}[noitemsep]
    \item \textbf{Ambiguity:} Combines \emph{Ambiguity} and \emph{Equivocation}.
    \item \textbf{Impossible Expectations}: Uses \emph{Impossible Expectations} and its alternative names.
    \item \textbf{False Equivalence}: Uses \emph{False Equivalence}.
    \item \textbf{False Dilemma}: Uses \emph{False Dilemma}.
    \item \textbf{Biased Sample Fallacy}: Uses \emph{Biased Sample Fallacy}.
    \item \textbf{Hasty Generalization}: Uses \emph{Hasty Generalization}.
    \item \textbf{Causal Simplification}: Combines \emph{False Cause Fallacy} and \emph{Single Cause Fallacy}.
    \item \textbf{Fallacy of Composition}: Combines \emph{Fallacy of Composition} and \emph{Fallacy of Division}.
    \item \textbf{Fallacy of Exclusion}: Uses \emph{Fallacy of Exclusion} and its alternative names.
\end{enumerate}

\section{Example Argument}
\label{appendix:example-argument}
\begin{figure*}
\small
    \centering
    \includegraphics[width=\textwidth]{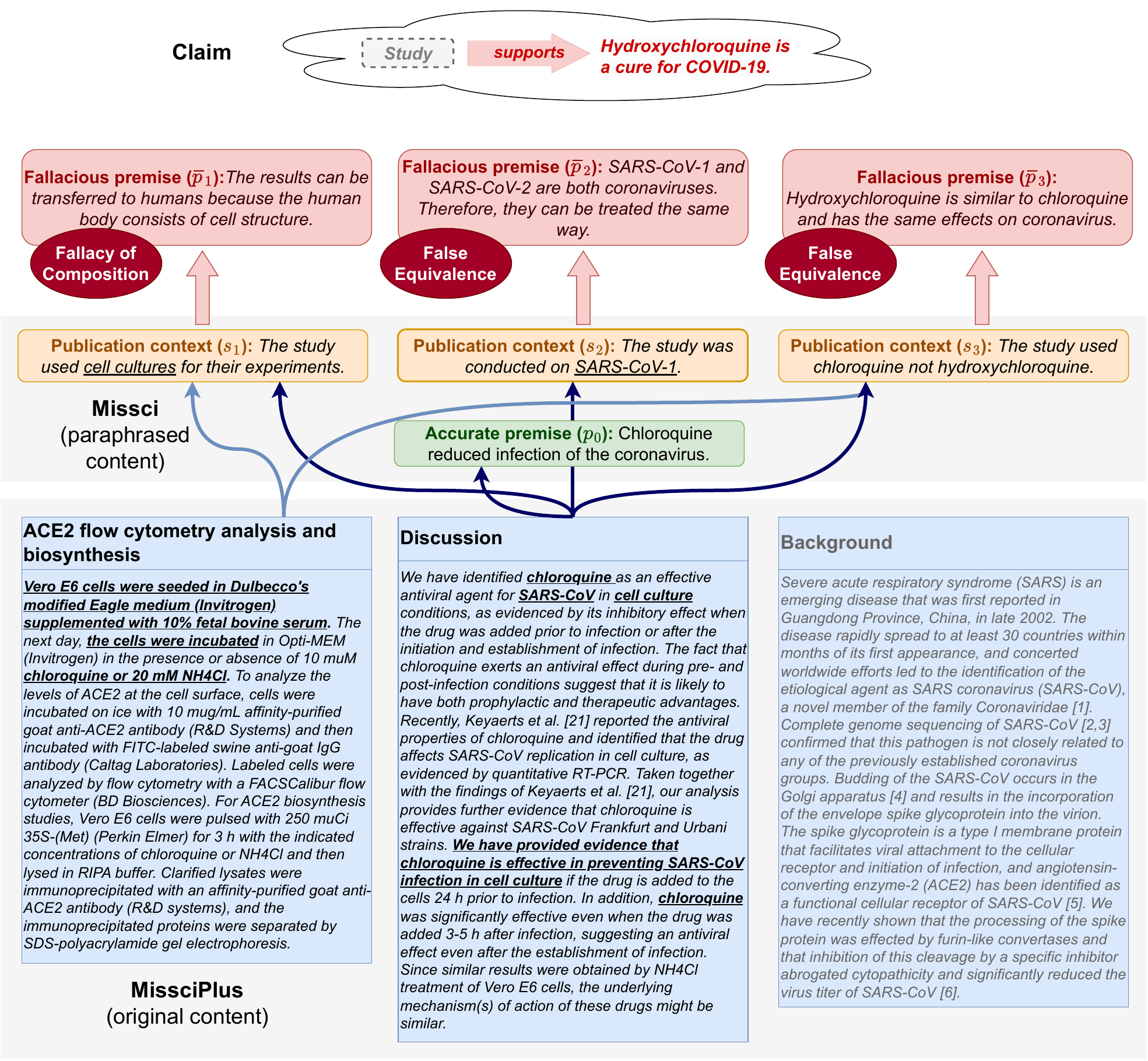}
    \caption{A complete fallacious argument, including three (out of 29) passages from the misrepresented study. The arrows indicate which \textcolor{blue}{\textbf{original passage}} in \dataset{} corresponds to the paraphrased \textcolor{orange}{\textbf{publication context}} in \missci{}. Some passages convey the same content in different publication contexts, and the same content may appear in multiple passages.}  
    \label{fig:figure-full-argument}
\end{figure*}

We present a complete fallacious argument, including three (out of 29) original passages (blue) from the misrepresented study, in Figure~\ref{fig:figure-full-argument}. The claim employs three distinct fallacies (red), each of which can be identified based on different information from the study. Two of the displayed passages contain relevant information for detecting these fallacies. The middle passage provides all the necessary information to identify all three fallacies, as well as the accurate premise. The first passage enables detection of the fallacies related to the study’s investigation being limited to cell cultures ($s_1$) and its use of chloroquine rather than hydroxychloroquine ($s_3$). In \dataset{}, only the original passages (marked in blue) are provided to the model, which must infer the fallacies from the complex scientific content. Although not used in this work, \dataset{} includes sentence-level annotations.

\section{Dataset Construction}
\label{appendix:dataset}

\subsection{Passage Extraction}
\label{appendix:dataset:step3:passage-selection:passage-extraction}
When documents are available in full-text via the PMC API\footnote{See \url{https://www.ncbi.nlm.nih.gov/books/NBK25499/\#chapter4.EFetch}}, we retrieve the XML document and collect text passages enclosed within \texttt{<p>} tags within the \texttt{<body>} node. If the full-text document is accessible as HTML but not through the API, we extract the HTML content from the Wayback Machine\footnote{\url{https://archive.org/web/}} and separate the passages based on HTML structure. We make the data collection script publicly available.

\subsection{Passage Selection}
\label{appendix:dataset:step3:passage-selection:passage-selection}
We segment each passage into constituent sentences using SciSpacy~\citep{neumann-etal-2019-scispacy} and compute the passage-level IMS~\citep{wright-etal-2022-modeling} by max-pooling the IMS of all sentences.
We retain the passage with the highest max-pooled IMS for each argument paraphrased information that needs a match ($p_0$, $s_i$). More passages (if available) are selected based on the highest IMS until we have six passages per argument. The IMS is particularly  well-suited for our problem as it quantifies the information of the scientific findings rather than focusing on semantic similarity

\subsection{Annotation}
\label{appendix:dataset:step3:annotations}
\begin{figure*}
\small
    \centering
    \includegraphics[width=\textwidth]{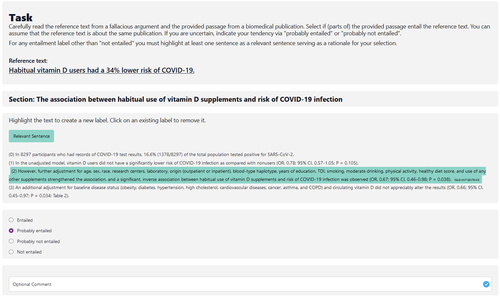}
    \caption{Annotation Interface from Surge AI to assign the entailment labels between the a passage and the paraphrased information.}  
    \label{fig:figure:annotation-mapping-interface}
\end{figure*}

\begin{table}[]
\small
    \centering
    \begin{tabular}{l |c c c c}
    \toprule
    \textbf{Stage} & \textbf{Args} & \textbf{Passages} & \textbf{Relations} & \textbf{Mapped} \\
    \midrule

    \textbf{1$^{\mathrm{st}}$ Round} & 118  & 681 & 2,191 & 64.3\% \\
    \textbf{2$^{\mathrm{nd}}$ Round} & 118  & 719  & 2,334 & 67.9\% \\
    \textbf{Consolid.} & 118  & 719  & 2,334 & 74.2\% \\
    \textbf{Cleaned} & 114 & 694  & 2,257  & 76.8\% \\

     \bottomrule
     
    \end{tabular}
    \caption{Stages of the argument mapping annotation, including how many of the paraphrased information are linked to passages. %
    }
    \label{tab:mapping-steps}
\end{table}

We use Surge AI\footnote{\url{https://www.surgehq.ai/}} as the annotation platform and employed two  M.Sc. students in biology paid 12.26 EUR per hour. We provide a screenshot of the annotation interface in Figure~\ref{fig:figure:annotation-mapping-interface} and construct \dataset{} in multiple rounds. Table~\ref{tab:mapping-steps} provides statistics for each round of dataset construction.

\paragraph{Passage Linking}
First, every combination of paraphrased information ($s_i$ or $p_0$) and selected passage $S_j$ is double annotated by our annotators, deciding whether (parts of) the passage entail the paraphrased information. To account for cases without a clear entailment label, we follow \citet{glockner2023ambifc} and allow annotators to express their uncertainty. Specifically, annotators can choose one of the labels ``entailed'', ``probably entailed'', ``probably not entailed'' or ``not entailed''.
We conservatively only consider the link between paraphrased information and passaged and as \emph{entailed} if at least one of both annotators labeled it as ``entailed'' and the other labeled it as either ``entailed'' or ``probably entailed''. In the first round, we annotated 2,191 relations with 681 distinct passages. This annotation round linked 266 paraphrased content to at least one passage.
In a second round, one annotator manually identified (if possible)  a corresponding passage from all remaining passages for each missing link.
These new passages were then double annotated with every paraphrased information ($s_i$, $p_0$) as outlined before. This annotation round contained 38 additional passages with 143 new relations and increased the number of paraphrased information linked to a passage to 281. The agreement for the fine-grained labels is 0.445. When merging ``entailed'' with ``probably entailed'' and ``not entailed'' with ``probably not entailed'', the inter-annotator agreement rises to 0.602. This suggests that some of the disagreement stems from uncertainty regarding a definite label. 

\paragraph{Consolidation}
To account for possible false negatives due to our conservative label aggregation, we consolidate annotations where the two annotators reached no unanimous label. We consider instances as not unanimous if they include instances labeled as ``probably entailed'' by both annotators or ``entailed'' by one and ``probably not entailed'' or ``not entailed'' by the other. To finalize the overall label, we provided the annotator with the entire fallacious argument during consolidation. This additional context allowed for a more contextual understanding of the role of the passage in reconstructing the fallacious argument. Our primary annotator, who had extensive experience in fallacy annotation and was involved in all our pilot studies, handled consolidation. Cases in which our consolidator could not clearly identify as ``entailed'' were labeled as ``not entailed''.
In total, 245 relations needed consolidation, 58 of which were consolidated as ``entailed'', leading to 26 previously unlinked paraphrased information.
Four arguments were not linked to any passages and were subsequently removed. This likely happened when the link to the misrepresented publication was falsely selected in \missci{}.

\subsection{Missing Passage Link Analysis}
\label{sec:dataset:missing-links}

\begin{table}[]
\scriptsize
    \centering
    \begin{tabularx}{\linewidth}{c|X}
    \toprule
    \textbf{\#} & \textbf{Sentence} \\
    \toprule
        
    \textbf{(1)} & The study did not include a group that did not wear masks at all. (\textit{negated})\\
    \midrule
    \textbf{(2)} & The experiments were done on concentrations that are different from concentrations found in patients or vaccinated people. (\emph{scope})\\
    \midrule
    \textbf{(3)} & Chloroquine diphosphate and hydroxychloroquine sulfate show antiviral activity against MERS-CoV and SARS-CoV. (\emph{multi-hop}) \\
   \midrule
    \textbf{(4)} & The average amount of spike protein in the blood was about 30 to 40 picograms/mL after receiving the Moderna vaccine. (\emph{multi-modal})\\

     \bottomrule
     
    \end{tabularx}
    \caption{Examples of paraphrased information with no linked passage.}
    \label{tab:no-mapping-examples}
\end{table} 

During dataset construction, we assumed that each paraphrased information from \missci{} could be linked to a single passage. This was feasible in 400 cases only (76.8\%). Accurate premises were more frequently linked to at least one passage (88.6\%), than publication contexts (72.0\%). We list representative reasons where no passages could not be found in Table~\ref{tab:no-mapping-examples}.
The most common reason, accounting for 41.6\%, was the presence of negation in the paraphrased information, discussing information \emph{not} present in the study. For instance, a claim questioning mask effectiveness would require a study with a control group without masks. However, if the study did not focus on mask effectiveness in general, there is no need for such a control group or to explicitly mention its absence. Negated sentences were more prevalent in undermining publication contexts (47.4\%) and less frequent in accurate premises (7.7\%).
Among the remaining instances, we identified (2) scope mismatches between the claim and the study, (3) information spread across multiple passages, and (4) non-textual components like tables or figures. While these fields could not be linked to a single passage, it is theoretically possible to reconstruct the argument using the complete publication.

\subsection{Licenses}
\missci{} extends the \missci{} dataset, published under the Apache 2.0 license.
\dataset{} aligns directly with their intended use, which is to outline the fallacious reasoning of distorted science transparently, and only improved its applicability in the real world. 
Our collected annotations and all scripts to collect and preprocess the scientific publications will be made publicly available under an open-source license. All fallacious logical argument and all publications in \dataset{} are in English.

\section{Length of Passages in \dataset{}}
\label{appendix:passage-length}
\begin{figure}
\small
    \centering
    \includegraphics[width=\linewidth]{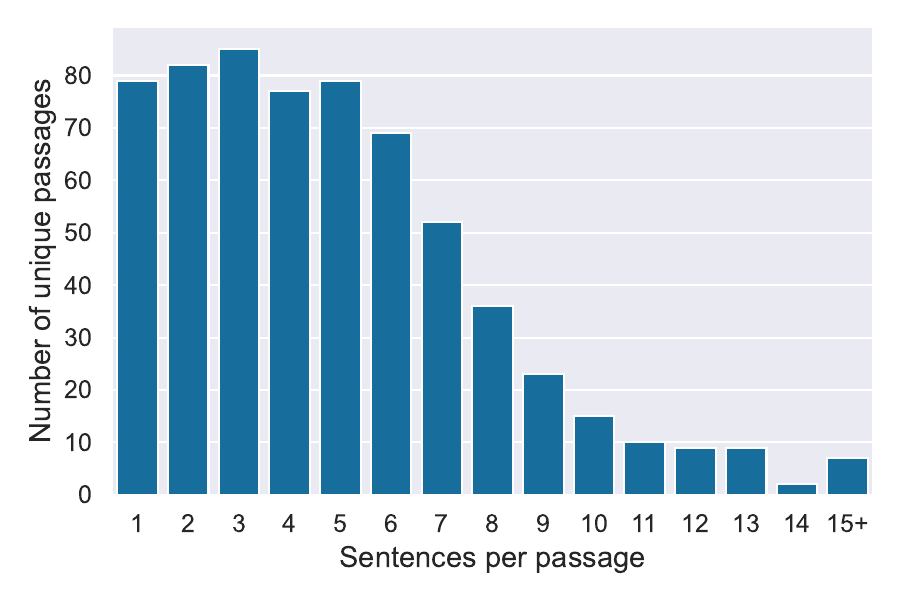}
    \caption{Number of sentences per passage over all \emph{annotated} passages in \dataset{}. Passages are only considered once (if used by multiple arguments).}  
    \label{fig:sentences-per-passage-selected}
\end{figure}

\begin{figure}
\small
    \centering
    \includegraphics[width=\linewidth]{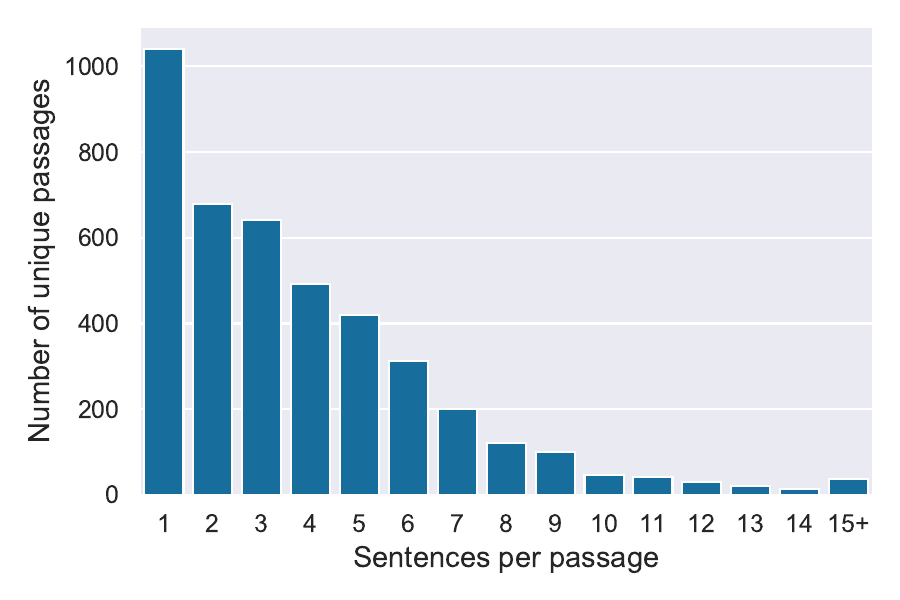}
    \caption{Number of sentences per passage over \emph{all} passages of the entire publication  in \dataset{}. Passages are only considered once (if used by multiple arguments).}  
    \label{fig:sentences-per-passage-all}
\end{figure}

The number of annotated passages per argument ranges from 1-8 passages (mean: 6.1; std: 1.2; median: 6.0). The number of all annotations per argument ranges from 1-114 passages (mean: 43.7; std: 25.5; median: 43.0).
Annotated passages vary in length, with 1 to 28 sentences. These passages contain 4.9 sentences on average (median: 4). When considering all passages of the entire misrepresented publication, each passage averages 3.8 sentences, with a median of 3. Figure~\ref{fig:sentences-per-passage-selected} displays the distribution of sentence counts in annotated passages, while Figure~\ref{fig:sentences-per-passage-all} shows the same for all passages.

\section{Scientific AFC Models}
\label{appendix:scientific-afc}

We train four DeBERTaV3~\citep{he2022debertav3} models for scientific AFC using \textsc{SciFact}~\citep{wadden-etal-2020-fact}, \textsc{CovidFact}~\citep{saakyan2021covidfact}, \textsc{HealthVer}~\citep{sarrouti-etal-2021-evidence-based} and the combination of all.

\subsection{Scientific AFC datasets}
\label{appendix:experiments:model-selection-afc:datasets}
\textsc{SciFact}~\citep{wadden-etal-2020-fact} covers multiple scientific domains using scientific abstracts as evidence documents. Citations were used to collect claims supported by the evidence. Refuted claims were generated via careful paraphrasing. While the original task includes the selection of evidence documents and rationale sentences, we only train models to predict the stance towards the claim given the full abstract as evidence. We used the official validation split as our test split (due to the hidden official test split). We randomly selected 200 instances from the official training set as the validation set and trained on the remaining instances.
\textsc{CovidFact}~\citep{saakyan2021covidfact} covers scientific claims related to COVID-19. Pairs of claims and evidence originate from a strictly moderated Reddit forum where every claim must provide an evidence document. Refuted claims were created by automatically changing the true claims. Unlike other AFC datasets, this dataset only distinguishes between ``supported'' and ``refuted'' claims, missing a label for ``not enough information''. 
\textsc{HealthVer}~\citep{sarrouti-etal-2021-evidence-based}
used COVID-19-related questions as queries and collected claims from resulting web pages via a search engine. Each claim was annotated against retrieved scientific abstracts as evidence using the three labels \textsc{Supports}, \textsc{Refutes}, and \textsc{Neutral}. 

\subsection{Hyperparameter search}
\label{appendix:experiments:model-selection-afc:models}
We fix the number of epochs to 5 and the seed to 1 and perform a hyperparameter search over the learning rates (1e-7, 5e-7, 1e-6, 5e-6, 1e-5, 5e-5) and batch sizes (4, 8, 16), evaluated on the respective validation set. When training on multiple datasets, the overall performance considers each instance equally (instead of averaging dataset scores). The results are shown in Tables~\ref{tab:hyperparameters-deberta-scifact}-\ref{tab:hyperparameters-deberta-sci-health-cov}.
We train five models using different seeds (1-5) based on the best-performing hyperparameters and report the averaged test performance in Table~\ref{tab:deberta-afc-test-results}.
\begin{table}[]
\small
    \centering
    \begin{tabular}{lcc|cc}
    \toprule
    \textbf{Model} & \textbf{lr} & \textbf{batch-size} & \textbf{Acc.} & \textbf{F1-score} \\
    \midrule
    deberta-v3-large & 1e-07 & 4 & 49.5 & 24.7 \\
    deberta-v3-large & 1e-07 & 8 & 49.5 & 24.7 \\
    deberta-v3-large & 1e-07 & 16 & 49.5 & 24.7 \\
    \midrule
    deberta-v3-large & 5e-07 & 4 & 50.5 & 26.2 \\
    deberta-v3-large & 5e-07 & 8 & 49.5 & 24.7 \\
    deberta-v3-large & 5e-07 & 16 & 49.0 & 22.1 \\
     \midrule
    deberta-v3-large & 1e-06 & 4 & 58.5 & 43.7 \\
    deberta-v3-large & 1e-06 & 8 & 53.0 & 31.6 \\
    deberta-v3-large & 1e-06 & 16 & 50.0 & 22.2 \\
     \midrule
    deberta-v3-large & 5e-06 & 4 & 90.5 & 88.6 \\
    deberta-v3-large & 5e-06 & 8 & 88.5 & 86.4 \\
    deberta-v3-large & 5e-06 & 16 & 82.5 & 79.6 \\
     \midrule
    deberta-v3-large & 1e-05 & 4 & \textbf{91.5} & \textbf{90.4} \\
    deberta-v3-large & 1e-05 & 8 & \textbf{91.5} & 90.2 \\
    deberta-v3-large & 1e-05 & 16 & 88.5 & 86.8 \\
     \midrule
    deberta-v3-large & 5e-05 & 4 & 50.0 & 22.2 \\
    deberta-v3-large & 5e-05 & 8 & 50.0 & 22.2 \\
    deberta-v3-large & 5e-05 & 16 & 82.5 & 80.7 \\
    \bottomrule
    \end{tabular}
    \caption{Hyperparameter search reported on our \textsc{SciFact} validation split. We fix the random seed to $1$ and number of epochs to $5$.
    }
    \label{tab:hyperparameters-deberta-scifact}
\end{table}

\begin{table}[]
\small
    \centering
    \begin{tabular}{lcc|cc}
    \toprule
    \textbf{Model} & \textbf{lr} & \textbf{batch-size} & \textbf{Acc.} & \textbf{F1-score} \\
    \midrule
    deberta-v3-large & 1e-07 & 4 & 69.0 & 40.8 \\
    deberta-v3-large & 1e-07 & 8 & 69.0 & 40.8 \\
    deberta-v3-large & 1e-07 & 16 & 69.0 & 40.8 \\
    \midrule
    deberta-v3-large & 5e-07 & 4 & 75.4 & 61.8 \\
    deberta-v3-large & 5e-07 & 8 & 69.0 & 40.8 \\
    deberta-v3-large & 5e-07 & 16 & 69.0 & 40.8 \\
    \midrule
    deberta-v3-large & 1e-06 & 4 & 85.0 & 83.2 \\
    deberta-v3-large & 1e-06 & 8 & 85.4 & 83.6 \\
    deberta-v3-large & 1e-06 & 16 & 69.0 & 40.8 \\
    \midrule
    deberta-v3-large & 5e-06 & 4 & 90.5 & 89.1 \\
    deberta-v3-large & 5e-06 & 8 & 90.2 & 88.8 \\
    deberta-v3-large & 5e-06 & 16 & 89.0 & 87.4 \\
    \midrule
    deberta-v3-large & 1e-05 & 4 & 88.8 & 87.4 \\
    deberta-v3-large & 1e-05 & 8 & \textbf{90.9} & \textbf{89.3} \\
    deberta-v3-large & 1e-05 & 16 & 88.8 & 87.1 \\
    \midrule
    deberta-v3-large & 5e-05 & 4 & 69.0 & 40.8 \\
    deberta-v3-large & 5e-05 & 8 & 69.0 & 40.8 \\
    deberta-v3-large & 5e-05 & 16 & 89.7 & 88.3 \\
    \bottomrule
    \end{tabular}
    \caption{Hyperparameter search reported on the \textsc{CovidFact} validation split. We fix the random seed to $1$ and number of epochs to $5$.}
    \label{tab:hyperparameters-deberta-covidfact}
\end{table}

\begin{table}[]
\small
    \centering
    \begin{tabular}{lcc|cc}
    \toprule
    \textbf{Model} & \textbf{lr} & \textbf{batch-size} & \textbf{Acc.} & \textbf{F1-score} \\
    \midrule
    deberta-v3-large & 1e-07 & 4 & 51.6 & 29.4 \\
    deberta-v3-large & 1e-07 & 8 & 49.8 & 29.3 \\
    deberta-v3-large & 1e-07 & 16 & 46.7 & 30.1 \\
    \midrule
    deberta-v3-large & 5e-07 & 4 & 72.3 & 65.6 \\
    deberta-v3-large & 5e-07 & 8 & 66.2 & 50.2 \\
    deberta-v3-large & 5e-07 & 16 & 64.8 & 47.5 \\
    \midrule
    deberta-v3-large & 1e-06 & 4 & 80.1 & 78.2 \\
    deberta-v3-large & 1e-06 & 8 & 76.5 & 72.0 \\
    deberta-v3-large & 1e-06 & 16 & 69.1 & 56.0 \\
    \midrule
    deberta-v3-large & 5e-06 & 4 & \textbf{86.6} & \textbf{84.6} \\
    deberta-v3-large & 5e-06 & 8 & 85.4 & 83.4 \\
    deberta-v3-large & 5e-06 & 16 & 86.3 & 84.5 \\
    \midrule
    deberta-v3-large & 1e-05 & 4 & 85.6 & 83.6 \\
    deberta-v3-large & 1e-05 & 8 & 85.7 & 83.4 \\
    deberta-v3-large & 1e-05 & 16 & 85.8 & 83.6 \\
    \midrule
    deberta-v3-large & 5e-05 & 4 & 51.8 & 22.7 \\
    deberta-v3-large & 5e-05 & 8 & 51.8 & 22.7 \\
    deberta-v3-large & 5e-05 & 16 & 51.8 & 22.7 \\
    \bottomrule
    \end{tabular}
    \caption{Hyperparameter search reported on the \textsc{HealthVer} validation split. We fix the random seed to $1$ and number of epochs to $5$.}
    \label{tab:hyperparameters-deberta-healthver}
\end{table}

\begin{table}[]
\small
    \centering
    \begin{tabular}{lcc|cc}
    \toprule
    \textbf{Model} & \textbf{lr} & \textbf{batch-size} & \textbf{Acc.} & \textbf{F1-score} \\
    \midrule
       deberta-v3-large & 1e-07 & 4 & 54.6 & 33.5 \\
    deberta-v3-large & 1e-07 & 8 & 52.9 & 28.1 \\
    deberta-v3-large & 1e-07 & 16 & 51.7 & 27.9 \\
    \midrule
    deberta-v3-large & 5e-07 & 4 & 69.6 & 62.5 \\
    deberta-v3-large & 5e-07 & 8 & 65.1 & 46.6 \\
    deberta-v3-large & 5e-07 & 16 & 64.0 & 46.0 \\
    \midrule
    deberta-v3-large & 1e-06 & 4 & 79.4 & 78.0 \\
    deberta-v3-large & 1e-06 & 8 & 76.5 & 73.6 \\
    deberta-v3-large & 1e-06 & 16 & 69.7 & 59.7 \\
    \midrule
    deberta-v3-large & 5e-06 & 4 & \textbf{85.2} & 83.6 \\
    deberta-v3-large & 5e-06 & 8 & 84.7 & 83.1 \\
    deberta-v3-large & 5e-06 & 16 & 84.9 & 83.6 \\
    \midrule
    deberta-v3-large & 1e-05 & 4 & 52.3 & 22.9 \\
    deberta-v3-large & 1e-05 & 8 & 84.8 & 83.1 \\
    deberta-v3-large & 1e-05 & 16 & 85.1 & \textbf{83.8} \\
    \midrule
    deberta-v3-large & 5e-05 & 4 & 52.3 & 22.9 \\
    deberta-v3-large & 5e-05 & 8 & 52.3 & 22.9 \\
    deberta-v3-large & 5e-05 & 16 & 52.3 & 22.9 \\
    \bottomrule
    \end{tabular}
    \caption{Hyperparameter search reported on the \textsc{SciFact}, \textsc{CovidFact}, and \textsc{HealthVer} validation split. We fix the random seed to $1$ and number of epochs to $5$.}
    \label{tab:hyperparameters-deberta-sci-health-cov}
\end{table}

\begin{table}[]
\small
    \centering
    \begin{tabular}{l|cc|ccc}
    \toprule
    & \multicolumn{2}{c|}{\emph{Overall}} & \multicolumn{3}{c}{\emph{Per Label (F1)}}\\
    \textbf{Datasets} & \textbf{Acc.} & \textbf{F1} & \textbf{S} & \textbf{N} & \textbf{R} \\
    \midrule
    \textsc{SciFact} & 88.9 & 87.9 & 92.0 & 86.0 &  85.8  \\
    \textsc{HealthVer} & 82.1 & 81.4 & 80.1 & 86.6 & 77.4\\
    \textsc{CovidFact} & 90.7 & 89.5 & 86.0 & -- & 93.1\\
    \midrule
   \emph{All} & 84.1 & 83.3 & 83.8 & 86.7 & 79.5\\
    \bottomrule
    \end{tabular}
    \caption{
    Evaluation of the scientific AFC models on the test set of the respective dataset used for training. The F1-score is macro averaged. The results are averaged over five seeds
}
    \label{tab:deberta-afc-test-results}
\end{table}

\section{Details on Finding the Kernel of Truth}
\label{appendix:subtask1:implementation}

\subsection{Implementation Details}
\subsubsection{Baselines}
As baselines, we randomly shuffle all passages (\emph{random}) using five different seeds (1-5) or order passages as they appear within the original publication (\emph{ordered}). We rank passages via BM25\footnote{\url{https://pypi.org/project/rank-bm25/}} for ranking based on lexical similarity. For pre-processing, we use SpaCy\footnote{\url{https://spacy.io/}} for tokenization and converting all tokens to lowercase. 

\subsubsection{Embedding-based approaches}
\label{appendix:st1:embedding}

    \begin{table}[]
    \small
    \centering
    \begin{tabular}{l|cccc}
    \toprule
    \textbf{Model} & \textbf{Agg.} & \textbf{Sent} & \textbf{MRR} & \textbf{P@1} \\
    \midrule
    BioBERT-ST & concat & all & 0.657 & 0.462 \\
BioBERT-ST & mean & 1 & 0.676 & \textbf{0.500} \\
BioBERT-ST  & mean & 2 & 0.664 & \textbf{0.500} \\
BioBERT-ST  & mean & 3 & \textbf{0.678} & \textbf{0.500} \\
\midrule
PubMedBERT-ST & concat & all & \textbf{0.764} & \textbf{0.654} \\
PubMedBERT-ST & mean & 1 & 0.620 & 0.423 \\
PubMedBERT-ST & mean & 2 & 0.661 &0.462 \\
PubMedBERT-ST & mean & 3 & 0.664 & 0.462 \\
\midrule
SapBERT-ST & concat & all & \textbf{0.701} & \textbf{0.538} \\
SapBERT-ST & mean & 1 & 0.664 & 0.500 \\
SapBERT-ST & mean & 2 & 0.680 & 0.462 \\
SapBERT-ST & mean & 3 &0.663 & 0.462 \\
\midrule
SBERT & concat & all & \textbf{0.663} & \textbf{0.500} \\
SBERT & mean & 1 &0.620 & 0.423 \\
SBERT & mean & 2 & 0.655 & 0.462 \\
SBERT & mean & 3 & 0.602 &0.385 \\
\midrule
SPICED IMS & concat & all & 0.699 & 0.538 \\
SPICED IMS & mean & 1 & 0.598 & 0.346 \\
SPICED IMS  & mean & 2 &\textbf{0.717} & \textbf{0.577} \\
SPICED IMS  & mean & 3 & 0.673 & 0.500 \\
\midrule
INSTRUCTOR & concat & all & 0.726 & 0.577 \\
INSTRUCTOR & mean & 1 & 0.671 & 0.500 \\
INSTRUCTOR & mean & 2 & \textbf{0.763} & \textbf{0.654} \\
INSTRUCTOR & mean & 3 & 0.747 & 0.615 \\
    \bottomrule
    \end{tabular}
    \caption{Hyper-parameter search on the validation split for sentence-embedding models to select passages $S_j^0 \Rightarrow p_0$.}
    \label{table:table-sentence-embedding-st1-dev}
\end{table} 

We experiment with various sentence-embedding models (in addition to those reported in Table~\ref{tab:results:subtask1-use}). This includes SBERT\footnote{\texttt{all-mpnet-base-v2}}~\citep{reimers-2019-sentence-bert}, and sentence transformers fine-tuned on BioBERT~\citep{lee2020biobert}, PubMedBERT~\citep{gu2021domain}, and SapBERT~\citep{liu-etal-2021-self} as provided by \citet{deka2022evidence}. For the INSTRUCTOR, we use the following prompts for the INSTRUCTOR~\citep{su2022one}, which follows the official templates:
\begin{itemize}[noitemsep]
    \item \textbf{Prompt (Claim):} \emph{``Represent the Scientific claim for retrieving supporting sentences: ''}
    \item \textbf{Prompt (Passage Sentences):} \emph{``Represent the Scientific sentence for retrieval: ''}
\end{itemize}
For all embedding models, we ranked the passages according to the cosine similarity to the embedded claim. For SPICED-IMS, we use the model\footnote{\url{https://github.com/copenlu/scientific-information-change}} provided by \citet{wright-etal-2022-modeling} and rank passages according to their IMS between claim and passage. We treat the scientific AFC models from §\ref{appendix:scientific-afc} as cross-encoder models to jointly encode each claim-passage pair and re-rank passages based on the predicted probability for the label \textsc{Supported}.

\paragraph{Hyperparameter Tuning}
We compare two perspectives on how to represent passages. First, the model embeds the entire passage by concatenating all sentences  (denoted as \emph{concat}). Second, we compute the cosine-similarity (or IMS) between the claim and each sentence of the passage individually and rerank passages based on the mean score of the top $k$ sentences with the highest score (denoted as \emph{mean}). The intuition is that only fractions of a passage may be relevant for the claim, and focusing on only parts of the passage can benefit the ranking performance. We report the performance on the validation set in Table~\ref{table:table-sentence-embedding-st1-dev} and select the best hyperparameters based on the P@$1$. 

\subsubsection{LLMs with PRP}
\label{appendix:subtask1:prp}

\begin{table*}[]
    \small
        \centering
        \begin{tabular}{lc| c|cccc}
        \toprule
        && &\multicolumn{4}{c}{\emph{MRR} after iteration \emph{i}} \\
        \textbf{Prompt} & \textbf{Section Title} & \textbf{P@1} & \textbf{i=1} & \textbf{i=2}  & \textbf{i=3} & \textbf{All} \\
        \midrule
        \citet{qin2023large} & \emph{no} & 0.628  & 0.750 & 0.778 & 0.783 & 0.783 \\
        \citet{qin2023large} (claim) & \emph{no} & 0.705  & 0.806 & 0.833 & 0.835 & 0.837 \\
        \citet{qin2023large} (convincing) & \emph{no} & 0.731  & 0.825 & 0.849 & 0.848 & 0.848 \\
        \citet{qin2023large} (support) & \emph{no} & 0.615  & 0.739 & 0.758 & 0.758 & 0.760 \\
        \midrule
        \citet{qin2023large} & \emph{yes} & 0.692  & 0.798 & 0.824 & 0.824 & 0.825 \\
        \citet{qin2023large} (claim) & \emph{yes} & 0.731  & 0.822 & 0.844 & 0.841 & 0.843 \\
        \citet{qin2023large} (convincing) & \emph{yes} & \textbf{0.744}  & \textbf{0.831} & \textbf{0.858} & \textbf{0.858} & \textbf{0.858} \\
        \citet{qin2023large} (support) & \emph{yes} &0.667  & 0.769 & 0.789 & 0.791 & 0.793 \\
        \bottomrule

    \end{tabular}
    \caption{ Evaluation of different prompts for PRP with Llama2-70B ( 8bit quantization), with random initialization. Averaged over three seeds.}
    \label{tab:results:subtask1:llm-pairwise-prompt-tuning}
\end{table*}

For PRP, we use the claim and two passages as input to the LLM and prompt it to output which passage should be ranked first. Following \citet{qin2023large}, we use these outputs to re-rank passages akin to bubble sort. The bubble sort algorithm ensures that the top $k$ elements are ranked at the top after $k$ iterations. 
We evaluate four prompts on the validation split in Table~\ref{tab:results:subtask1:llm-pairwise-prompt-tuning}, starting with the prompt provided by \citet{qin2023large} and three deviations tailored towards our task. We further assess whether passages should include or exclude the title and found that including the title generally improves performance. During prompt selection in Table~\ref{tab:results:subtask1:llm-pairwise-prompt-tuning} we randomly shuffle all passages prior to reranking them to avoid the impact of the strong positional bias on the validation split (P@$1$: 0.577).  Since our metrics are only sensitive to the top-ranked results, PRP can only be completed over a few iterations to reduce costs without affecting the performance much. Following our experiments on the validation split, we only ran PRP for three iterations on the test set.
Prompts are listed in §\ref{appendix:prp-prompts}.

\subsection{Exhaustive Results on \dataset{}}
\begin{table*}[]
\small
    \centering
    \begin{tabular}{l | cc | ccc}
    \toprule
    & \multicolumn{2}{c|}{Annotated Passages (\emph{closed})} &  \multicolumn{3}{c}{All Passages \emph{(open)}} \\ 
    \textbf{Model} & \textbf{P@1} & \textbf{MRR}   & \textbf{P@1} & \textbf{MRR} & \textbf{HasPos@3}\\
    \toprule
    \emph{Random} & 0.360 & 0.566  & 0.096 & 0.209 & 0.213\\
    \emph{Ordered} & 0.480 & 0.658  & 0.320 & 0.443 & 0.507\\
    \midrule
    BM25 & 0.547 & 0.705  & 0.387 & 0.539 & 0.627\\
    \midrule
    SBERT~\citep{reimers-2019-sentence-bert} & 0.400 & 0.631  & 0.280 & 0.460 & 0.547\\

    PubMedBERT ST~\citep{deka2022evidence} & 0.440 & 0.652  & 0.240 & 0.442 & 0.573\\
    BioBERT ST~\citep{deka2022evidence} & 0.547 & 0.712  & 0.427 & 0.582 & 0.680\\
    SapBERT ST~\citep{deka2022evidence} & 0.480 & 0.672  & 0.333 & 0.514 & 0.627\\

    INSTRUCTOR~\citep{su2022one}  & 0.573 & 0.738  & 0.480 & 0.631 & 0.733\\
    SPICED-IMS \citep{wright-etal-2022-modeling} & 0.587 & 0.742  & \textbf{0.533} & \textbf{0.664} & \textbf{0.760}\\
    
    \midrule
    DeBERTa$_{v3}$ SciFact \citep{wadden-etal-2020-fact} & 0.603 & 0.748  & 0.389 & 0.535 & 0.627\\
    DeBERTa$_{v3}$ CovidFact \citep{saakyan2021covidfact} & 0.517 & 0.691  & 0.307 & 0.450 & 0.507\\
    DeBERTa$_{v3}$ HealthVer \citep{sarrouti-etal-2021-evidence-based}&  0.608 & 0.765  & 0.347 & 0.516 & 0.629\\
    DeBERTa$_{v3}$ Scienfic AFC (\emph{all}) & 0.608 & 0.768  & 0.349 & 0.514 & 0.600\\

    \midrule
    Llama2-70B \citep{touvron2023llama} PRP (it=3) & 0.711 & 0.830  & -- & -- & --\\

    Llama3-8B PRP (it=3) & 0.729 & \textbf{0.850}  & -- & -- & --\\
GPT3.5 PRP (it=3) & 0.671 & 0.815  & -- & -- & --\\
    GPT4 \citep{openai2023gpt4} PRP (it=3) & \textbf{0.742} & \textbf{0.850}  & -- & -- & --\\
     \bottomrule
     
    \end{tabular}
    \caption{Ranking performance to find the passages based on which the claim was made ($S_j^0$) over all 75 test instances where $p_0$ was linked to at least one passage.}
    \label{tab:results:subtask1-complete}
\end{table*} 

Table~\ref{tab:results:subtask1-complete} provides a list of all evaluated models, including additional sentence embedding models, tasked to find the passage $S^0$ based on which the claim was made. We additionally report HasPositives@$3$ \citep{shaar-etal-2020-known} as a more relaxed measure that allows an appropriate passage to be ranked within the top three results.

\section{Details on Finding Undermining Passages}
\label{appendix:subtask2}
We only evaluate finding undermining passages in the open subset. Hence, the results represent a lower bound of the system's performance. Per our hyper-parameter search (§\ref{appendix:subtask2:afc-ablation-closed} and §\ref{appendix:subtask2:embedding-ablation-open}), we always provide one randomly sampled passage $S^0$ to the model, which is necessary to understand the rationale (and reasoning gaps) behind the claim. If no passage $S^0$ exists, we use the paraphrased accurate premise from \missci{} instead. All experiments are averaged over five seeds (1-5).

\subsection{Implementation Details}
\subsubsection{AFC as Rankers}
\label{appendix:subtask2:afc-ablation-closed}

\begin{table}[]
\small
    \centering
    \begin{tabular}{l c | c cc}
    \toprule
    && \multicolumn{3}{c}{\emph{mAP when $S_j^0$ included as}} \\
    \textbf{Train data} & \textbf{Strategy} & \textbf{No} $S_j^0$ & \textbf{Claim} & \textbf{Evid.} \\
    \midrule
    SciFact & \emph{Support} & 0.304 & 0.389 & 0.193 \\
    SciFact & \emph{Refute} &0.258 & 0.251 & 0.230 \\
    SciFact & \emph{Both} &0.348 & \textbf{0.404} & 0.208 \\
    \midrule
    CovidFact & \emph{Support} & 0.236 & \textbf{0.369} & 0.206 \\
    CovidFact & \emph{Refute} & 0.161 & 0.150 & 0.214 \\
    \midrule
    HealthVer & \emph{Support} & 0.270 & 0.357 & 0.235 \\
    HealthVer & \emph{Refute} &0.234 & 0.269 & 0.241 \\
    HealthVer & \emph{Both} & 0.273 & \textbf{0.375} & 0.245 \\
    \midrule
    \emph{All} AFC & \emph{Support} & 0.308 & 0.410 & 0.188 \\
    \emph{All} AFC & \emph{Refute} & 0.270 & 0.183 & 0.192 \\
    \emph{All} AFC & \emph{Both} &0.355 &  \textbf{0.420} & 0.191 \\

    \bottomrule
    \end{tabular}
    \caption{Evaluation of ranking passages based on the predicted probability of \emph{supported}, \emph{refuted} or their sum \emph{(both)} on the validation split.  }
    \label{tab:results:subtask2:scifact:threshold}
\end{table} 

A key challenge in retrieving passages that indicate reasoning gaps in a zero-shot setting is the absence of directly comparable tasks. For instance, passages containing content that undermines an argument may convey a supporting stance (e.g., if the claim is confirmed based on a small sample size, as in \emph{Hasty Generalization}). When re-purposing AFC models to predict whether $S_j$ points to a reasoning gap in the argument, we experiment with various strategies that aggregate the stance prediction labels from AFC models. Specifically, we measure the predicted probability mass for the labels \textsc{Supported}, \textsc{Refuted}, or their sum\footnote{Except when trained on CovidFact, which only exhibits two labels.} (\emph{both}). 
For each strategy, we experiment with the role of the passage $S^0$ based on which the claim was made:
\begin{itemize}[noitemsep]
    \item \textbf{No $S^0$:} The AFC model sees only $\overline{c}$ and $S_j$.
    \item \textbf{Claim:} The claim is reformulated as an argument by incorporating $S^0$ through the expression ``$S^0$ \texttt{Therefore:} \emph{claim}''.
    \item \textbf{Evid.:} $S^0$ is added at the beginning of the evidence passage ($S_j$) that is subject to ranking.
\end{itemize}
Table~\ref{tab:results:subtask2:scifact:threshold} shows the validation split results. We select the best-performing strategy according to MAP, which always concatenates a randomly sampled passage $S^0$ with the claim.

\subsubsection{Dense embedding rankers}
\label{appendix:subtask2:embedding-ablation-open}
\begin{table}[]
\small
    \centering
    \begin{tabular}{l|cc}
    \toprule
    & \multicolumn{2}{c}{\textbf{MAP}} \\
    \textbf{Model} & \emph{no} $S_j^0$ &  $S_j^0$  \\
    \toprule

    BM25  & 0.283 &  \textbf{0.554} \\
    \midrule
    SBERT (\emph{concat}) & 0.328 & \textbf{0.522} \\
    SBERT (\emph{mean-1}) & 0.333 & 0.366 \\
    SBERT (\emph{mean-2}) &0.353 & 0.354 \\
    SBERT (\emph{mean-3}) & 0.308 & 0.342 \\
    \midrule
    INSTRUCTOR (\emph{concat}) & 0.388 & \textbf{0.582} \\
    INSTRUCTOR (\emph{mean-1}) &0.404 & 0.480 \\
    INSTRUCTOR (\emph{mean-2}) &0.389 & 0.516 \\
    INSTRUCTOR (\emph{mean-3}) & 0.368 & 0.479 \\
    \midrule
    SPICED-IMS (\emph{concat}) &0.384 & \textbf{0.562} \\
    SPICED-IMS (\emph{mean-1}) & 0.379 & 0.429 \\
    SPICED-IMS (\emph{mean-2}) & 0.380 & 0.408 \\
    SPICED-IMS (\emph{mean-3}) &0.323 & 0.380 \\
    \midrule
    PubMedBERT ST (\emph{concat}) &0.283 & 0.490 \\
    PubMedBERT ST (\emph{mean-1}) & 0.337 & \textbf{0.529} \\
    PubMedBERT ST (\emph{mean-2}) & 0.318 & 0.513 \\
    PubMedBERT ST (\emph{mean-3}) &0.324 & 0.490 \\
    \midrule
    BioBERT ST (\emph{concat}) & 0.270 & 0.516 \\
    BioBERT ST (\emph{mean-1}) & 0.313 & 0.531 \\
    BioBERT ST (\emph{mean-2}) & 0.311 & \textbf{0.536} \\
    BioBERT ST (\emph{mean-3}) & 0.280 & 0.521 \\
    \midrule
    SapBERT ST (\emph{concat}) & 0.323 & 0.484 \\
    SapBERT ST (\emph{mean-1}) & 0.312 & 0.497 \\
    SapBERT ST (\emph{mean-2}) & 0.325 & \textbf{0.500} \\ %
    SapBERT ST (\emph{mean-3}) & 0.323 & \textbf{0.500} \\
    \bottomrule
    \end{tabular}
    \caption{ Ranking performance of embedding-based models on the validation instances to select passages linked to fallacies. We select the best model based on the highest MAP.}
    \label{tab:ablation:subtask2:embedding:p0}
\end{table}

Using the same ranking models via embeddings or BM25 as in §\ref{appendix:subtask1:implementation}, we assess the impact of prepending a passage $S^0$ to the claim on the validation split in Table~\ref{tab:ablation:subtask2:embedding:p0}. We select the best configuration according to mAP for the test split. For the INSTRUCTOR, we modify the prompt to encode the claim by looking for ``refuting'' instead of ``supporting'' sentences (cf. §\ref{appendix:st1:embedding}), which resulted in a higher MAP on the validation set.

\subsection{Impact of $S^0_j$ Passages}
\label{appendix:subtask2:fallacy-analysis}
\begin{figure}
\small
    \centering
    \includegraphics[width=\linewidth]{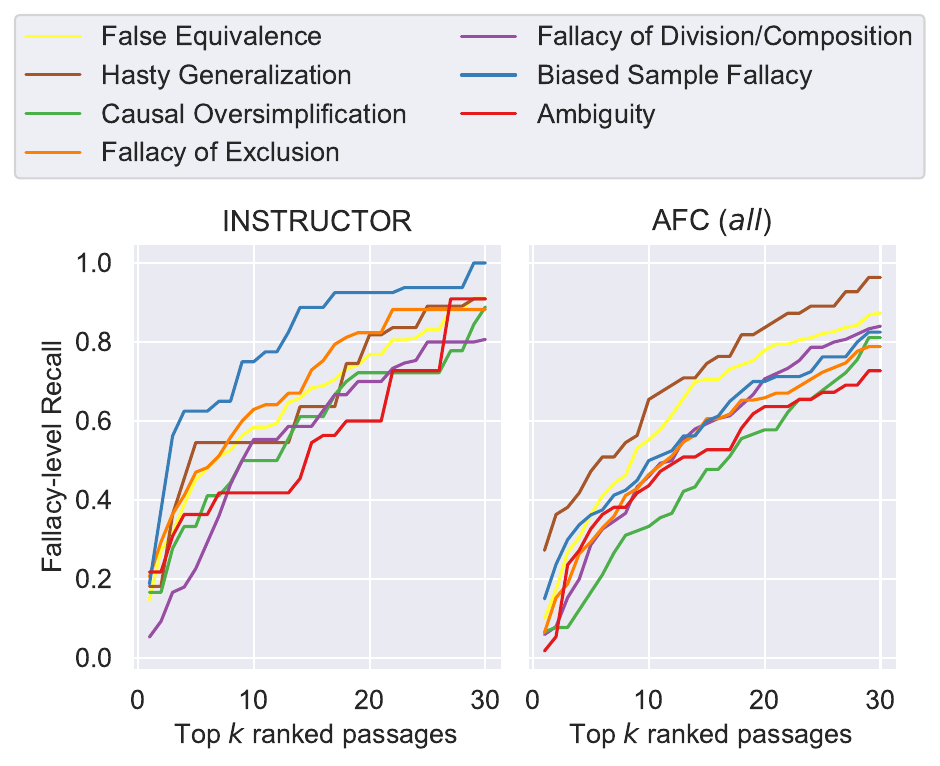}
    \caption{Recall of undermining passages per fallacy class over the top $k$ ranked passages when removing all $S^0$ passages (based on which the claim was made). Only listing fallacies with $\geq 20$ occurrences.}  
    \label{fig:figure-st2-afc-instructor-per-fallacy-nop0}
\end{figure}

We artificially remove all passages from the ranked results if the passage communicates the content based on which the claim was made (i.e., $S_j^0$ passages) in Figure~\ref{fig:figure-st2-afc-instructor-per-fallacy-nop0}. Among these passages only, both rankers exhibit different strengths. INSTRUCTOR prefers passages based on which the \emph{Biased Sample Fallacy} or the \emph{Fallacy of Exclusion} can be detected. The AFC ranker prioritizes fallacies based on which \emph{Hasty Generalization} and \emph{False Equivalence} can be detected. An intuitive explanation is that both fallacies are similar to the support relationship and, hence, to the predicted probability for the label \textsc{Supported}.

\subsection{Exhaustive Results on \dataset{}}
\begin{table*}[]
\small
    \centering
    \begin{tabular}{l|c|ccc | ccc}
    \toprule
    & \multicolumn{4}{c|}{\emph{Passage-wise}} & \multicolumn{3}{c}{\emph{Fallacy-wise}} \\
    \textbf{Model} & \textbf{mAP} & \textbf{P@1} & \textbf{P@3} & \textbf{P@10}  & \textbf{R@1} & \textbf{R@3}& \textbf{R@10} \\
    \toprule

    \emph{Random}  & 0.205 & 0.136 & 0.130 & 0.124 & 0.168 & 0.252 & 0.438\\
    \emph{Ordered}  & 0.286 & 0.298 & 0.210 & 0.145 & 0.266 & 0.374 & 0.497\\
    BM25  & 0.496 & 0.617 & 0.347 & 0.212 & 0.413 & 0.485 & 0.602\\
    \midrule
    SBERT~\citep{reimers-2019-sentence-bert}   & 0.520 & 0.640 & 0.381 & \textbf{0.221} & 0.423 & 0.503 & 0.609\\
    INSTRUCTOR~\citep{su2022one}   & \textbf{0.541} & \textbf{0.652} & \textbf{0.409} & \textbf{0.222} & \textbf{0.439} & 0.529 & \textbf{0.613}\\
    SPICED-IMS~\citep{wright-etal-2022-modeling}  & 0.524 & 0.640 & 0.388 & \textbf{0.219} & 0.423 & 0.516 & 0.595\\
    PubMedBERT ST~\citep{deka2022evidence}  & 0.489 & 0.588 & 0.360 & 0.207 & 0.387 & \textbf{0.548} & 0.605\\
    BioBERT ST~\citep{deka2022evidence}  & 0.491 & 0.600 & 0.377 & 0.200 & 0.400 & 0.495 & 0.570\\
    SapBERT ST~\citep{deka2022evidence}   & 0.504 & 0.619 & 0.379 & 0.211 & 0.411 & 0.547 & \textbf{0.615}\\
    
    \midrule
    DeBERTaV3 SciFact \citep{wadden-etal-2020-fact}   & 0.360 & 0.326 & 0.266 & 0.172 & 0.264 & 0.393 & 0.518\\
    DeBERTaV3 CovidFact \citep{saakyan2021covidfact}   & 0.380 & 0.457 & 0.260 & 0.165 & 0.326 & 0.407 & 0.538\\
    DeBERTaV3 HealthVer \citep{sarrouti-etal-2021-evidence-based}  & 0.368 & 0.410 & 0.284 & 0.176 & 0.321 & 0.426 & 0.544\\
    DeBERTaV3 Scientific AFC (\emph{all})   & 0.306 & 0.338 & 0.263 & 0.177 & 0.267 & 0.400 & 0.554\\
    \bottomrule
    \end{tabular}
    \caption{Ranking results for finding passages linked to fallacies on the test split, using  all passages of the misrepresented publication.}
    \label{tab:results:subtask2-open-complete}
\end{table*} 

We list all results of detecting undermining passages on the \dataset{} test split in Table~\ref{tab:results:subtask2-open-complete}.

\section{Details on Argument Reconstruction}
\label{appendix:subtask3}

\subsection{LLM argument reconstruction}
\label{appendix:llm-argument-reconstruction-details}

\begin{table*}[]
\small
    \centering
    \begin{tabular}{lc|ccc|cc|c}
    \toprule
    &&\multicolumn{3}{c|}{\emph{Fallacy Level (premise+class)}} & \multicolumn{2}{c|}{\emph{Arg. Level}} & \emph{Count}\\
    \textbf{Prompt} & \textbf{Passages} & \textbf{R@5} & \textbf{P@5} & \textbf{P@1}& \textbf{F1@5} & \textbf{Arg@1} & \textbf{Pred/Arg}  \\
    \midrule
    p1-basic & per-passage & 0.079 & 0.046 & 0.133 & 0.306 & 0.167 & 2.6 \\
p2-support & per-passage & 0.083 & 0.048 & 0.078 & 0.296 & 0.200 & 2.7 \\
p3-undermine & per-passage & 0.079 & 0.047 & 0.089 & 0.302 & 0.178 & 2.4 \\
p4-connect & per-passage & 0.176 & \textbf{0.134} & \textbf{0.222} & 0.359 & 0.333 & 1.4 \\
p5-auto & per-passage & 0.083 & 0.066 & 0.144 & 0.263 & 0.178 & 1.4 \\
p6-auto-connect & per-passage & 0.111 & 0.100 & 0.156 & 0.272 & 0.256 & 1.1 \\
p7-passage & per-passage & 0.093 & 0.047 & 0.078 & 0.357 & 0.222 & 4.2 \\
p8-passage-assumptions & per-passage & 0.162 & 0.084 & 0.122 & 0.379 & 0.356 & 4.2 \\
p9-passage-detective & per-passage & 0.157 & 0.146 & 0.222 & 0.270 & 0.344 & 1.1 \\
p10-passage-evaluate & per-passage & 0.120 & 0.060 & 0.111 & 0.337 & 0.289 & 4.8 \\
p11-passage-evaluate-2 & per-passage & 0.176 & 0.087 & 0.178 & 0.353 & 0.367 & 4.6 \\
p12-passage-detective-2 & per-passage & \textbf{0.218} & 0.113 & 0.211 & \textbf{0.360} & \textbf{0.444} & 3.9 \\
p13-passage-evaluate-3 & per-passage & 0.167 & 0.083 & 0.167 & 0.359 & 0.300 & 4.4 \\
\midrule
p1-basic & all passages & 0.060 & 0.053 & 0.067 & 0.294 & 0.144 & 2.8 \\
p2-support & all passages & 0.065 & 0.058 & 0.044 & 0.303 & 0.156 & 2.8 \\
p3-undermine & all passages & 0.046 & 0.040 & 0.044 & 0.280 & 0.111 & 2.9 \\
p4-connect & all passages & 0.093 & \textbf{0.162} & \textbf{0.178} & 0.239 & 0.211 & 1.4 \\
p5-auto & all passages & 0.056 & 0.070 & 0.089 & 0.247 & 0.133 & 2.0 \\
p6-auto-connect & all passages & 0.046 & 0.110 & 0.111 & 0.163 & 0.111 & 1.0 \\
p7-passage & all passages & 0.042 & 0.024 & 0.022 & 0.333 & 0.100 & 4.4 \\
p8-passage-assumptions & all passages & 0.144 & 0.088 & 0.089 & 0.376 & 0.289 & 4.2 \\
p9-passage-detective & all passages & 0.069 & 0.153 & 0.133 & 0.218 & 0.167 & 1.1 \\
p10-passage-evaluate & all passages & 0.130 & 0.069 & 0.167 & 0.377 & 0.233 & 4.7 \\
p11-passage-evaluate-2 & all passages & 0.144 & 0.077 & 0.089 & 0.371 & 0.256 & 5.1 \\
p12-passage-detective-2 & all passages & \textbf{0.171} & 0.105 & 0.144 & 0.379 & 0.400 & 3.9 \\
p13-passage-evaluate-3 & all passages & 0.162 & 0.094 & 0.124 & \textbf{0.402} & \textbf{0.344} & 4.4 \\

    \bottomrule

    \end{tabular}
    \caption{Argument reconstruction prompt-tuning using Llama3-8B-Instruct on the validation split. }
    \label{tab:subtask3-prompt-engneering-dle-llama3:dev}
\end{table*} 

We provide the LLM with all passages $S_j$ that were linked to at least one reasoning gap (or fallacy), and one passage $S^0$ that communicates the content based on which the claim was made. If multiple candidates for $S^0$ exist, we randomly select one. Experiments over \missci{} use the paraphrased $p_0$ instead of a randomly sampled passage $S^0$, and the publication context $s_i$ instead of the linked passages $S_j$. We run each experiment over three different seeds (1-3). For Llama3-8B we use a temperature of 0.3. For a fair assessment given the changed requirements for the LLM, we perform extensive prompt-search over all six prompts evaluated for \missci{} and an additional new prompts (cf. §\ref{appendix:reconstruction-prompts}). During prompt search, we provide the complete fallacy information consisting of the definition, the logical form, and an example (cf. §\ref{appendix:fallacies}; Tables~\ref{tab:fallacy-overview}-\ref{tab:fallacy-overview-examples}) within the instructions. All prompts task LLMs to output a ranked list of verbalized fallacious premises and the applied fallacy class. If possible, fallacies outside our inventory were converted; otherwise, they were removed. The results are listed in Table~\ref{tab:subtask3-prompt-engneering-dle-llama3:dev}. We additionally report the average number of predicted fallacies per argument. This number excludes fallacies that the LLM hallucinated and are invalid (e.g., \emph{Contextomizer}, \emph{Accident}, \emph{Conflict of Interest}, \emph{Fallacy of Conclusion}) or that are outside of our fallacy inventory (e.g., \emph{Fear Mongering}, \emph{Non Sequitur}, \emph{Ad Hominem}, \emph{False Consensus}).

\subsection{Llama3 Judge $\phi$}
\label{subtask3-judge}

\begin{table}[]
\small
    \centering
    \begin{tabular}{l c c| ccc}
    \toprule
    \textbf{Method} & \textbf{Prompt} & \textbf{Extra} & \textbf{F1} &  \textbf{Acc.} \\
    \midrule
    \emph{Majority} & -- & -- & 0.372 & 0.593 \\
    NLI-S + LR & -- & -- & 0.642 & 0.695 \\
    \midrule
    Zeroshot & p1 & 0-shot & 0.570 & 0.581 \\
    Zeroshot  & p2 & 0-shot & 0.636 & 0.647 \\
    Zeroshot  & p3 & 0-shot & 0.636 & 0.641 \\
    Zeroshot  & p4 & 0-shot & 0.626 & 0.629 \\
    Zeroshot  & p5 & 0-shot & 0.562 & 0.569 \\
    \midrule
    ICL & p1 & 8-shot & 0.623 & 0.625 \\
    ICL & p2 & 8-shot & 0.635 & 0.639 \\
    ICL & p3 & 8-shot & 0.620 & 0.623 \\
    ICL & p4 & 8-shot & 0.604 & 0.607 \\
    ICL & p5 & 8-shot & 0.612 & 0.617 \\
    \midrule
    ICL & p1 & 16-shot & 0.656 & 0.663 \\
    ICL & p2 & 16-shot & 0.640 & 0.645 \\
    ICL & p3 & 16-shot & 0.645 & 0.651 \\
    ICL & p4 & 16-shot & 0.656 & 0.661 \\
    ICL & p5 & 16-shot & 0.652 & 0.657 \\
    \midrule
    SFT & p4 & 1-epochs & 0.671 & 0.687 \\
    SFT & p4 & 2-epochs & 0.690 & 0.711 \\
    SFT & p4 & 3-epochs  & 0.725 & 0.749 \\
    SFT & p4 & 4-epochs & 0.751 & 0.770 \\
    SFT & p4 & 5-epochs & \textbf{0.788} & \textbf{0.798} \\
    SFT & p4 & 6-epochs  & 0.761 & 0.776 \\

    \bottomrule
    \end{tabular}
    \caption{Cross-validation evaluation for implementations of the $\phi^{\mathrm{p+f}}$ judge. All prompt-based approaches use Llama3-8B-Instruct as backend.}
    \label{tab:llm-judge-llama3}
\end{table} 

We experiment with four different prompts (cf. §\ref{appendix:judge-prompts}) with the Llama3-8B-Instruct model as a binary classifier $\phi$ that determines whether two fallacious premises exhibit fallacious reasoning that bridges the identical gap. We use the human evaluation data provided by \citet{glockner2024dismantling} for training data. We discard all trivially invalid instances where the generated premise $\hat{\overline{p}}_i$ was almost identical to the claim or the paraphrased publication context $s_i$ by discarding all premise pairs if 
\[
\mathrm{min}\Bigl[\mathrm{lev(\hat{\overline{p}}_i, \overline{c})}, \mathrm{lev(\hat{\overline{p}}_i, \overline{s}_i)}\Bigr] \leq t
\]
where $\mathrm{lev}$ is the Levensthein distance and the threshold $t=2$.
This yielded a total of 168/240 manually annotated premise pairs.
To adapt Llama3-8B-Instruct for the task, we perform (1) zero-shot experiments, (2) in-context learning \citep{brown2020language} (ICL) experiments and (3) supervised fine-tuning (SFT) using QLoRA~\citep{dettmers2023qlora} and 8bit quantization \citep{dettmers2022bit}.
We validate each approach using five-fold cross-validation where folds are separated by the arguments, ensuring the LLM is evaluated with premises from unseen arguments. We set the temperature to zero and evaluate all ICL and SFT experiments over three random seeds ($1$,$2$,$3$) to account for different (ordering) of seen instances.
The results are listed in Table~\ref{tab:llm-judge-llama3}. 
As a baseline, we report the performance of always predicting the \emph{majority} label. We further report a baseline using a univariate logistic regression (\emph{LR}) on top of the automatic \emph{NLI-S}~\citep{glockner2024dismantling}, which showed the highest correlation with human judgment in \missci{}. NLI-S uses the predicted entailment probability of a T5~\citep{raffel2020exploring} model fine-tuned by \citet{honovich-etal-2022-true}. 
Rather than only considering the entailment score given the reference text as a premise and generated text as a hypothesis, NLI-S swaps its roles to avoid penalizing the model if it generates more specific text.
Without SFT, ICL with 16 shots and the template \emph{p4} reached the best performance measured via the F1-score. We consequentially performed SFT using the same prompt, which led to the overall best model with fine-tuning for $5$ epochs with a \emph{linear} schedule, a learning rate of $5e-4$, a batch-size of $4$ and $\alpha=16$, $r=64$, $\mathrm{dropout}=0.2$ for QLoRA. We use these hyperparameters to fine-tune Llama3-8B-Instruct on the entire data and use it as the backend model for $\phi^\mathrm{f+p}$.

\subsubsection{Prompting Strategies}
Prompts using \emph{per-passage} prompting follow the prompting scheme of \missci{} and have a dedicated field for the content based on which the claim was made ($p_0$ or $S_j^0$) and for the publication context necessary to detect the fallacy ($s_i$ or $S_j$). This prompting technique requires $n$ separate prompts for $n$ passages. Specifically, we create $n-1$ prompts using a randomly sampled passage $S^0$ (based on which the claim was made) together with each other passage linked to a fallacy, separately. In \dataset{}, the passage $S^0$ itself may be linked to fallacies. Hence, we prompt the model again with only the passage $S^0$. When selecting the top $k$ results of multiple prompts for the same argument, combine all results, consisting of the fallacious premise and fallacy class, while keeping their ranking information. We then return the top $k$ results based on their prompt-specific rank. We prefer results with different fallacy classes when multiple fallacies share the same rank to avoid sampling the same fallacious reasoning numerous times.
Prompts that concatenate \emph{all passages} within a single prompt follow the holistic view of arguments. We sort all passages based on the order in which they occur in the scientific publication and only prompt the LLM once per argument. 

\subsubsection{Performance on \missci{} with only linked publication context}
\label{appendix:sec:missci_onlymapped}

\begin{table*}[]
\small
    \centering
    \begin{tabular}{ll|ccc} \\
    \toprule
      \textbf{LLM} & \textbf{Info} & \textbf{R@5} ($\phi^{\mathrm{f+p}}$) & \textbf{R@5} ($\phi^{\mathrm{f}}$) &\textbf{Arg@$1$ ($\phi^{\mathrm{f+p}}$)} \\
      \midrule
      & DLE & 0.255 & 0.520 & 0.516  \\
      \multirow{ 2}{*}{Llama3-8B} & DL & 0.237 & 0.453 & 0.520  \\
       & DE & 0.220 & 0.470 & 0.468  \\
        & LE & 0.251 & 0.492 & 0.504  \\
        \midrule
    & DLE & 0.217 & 0.456 & 0.464  \\
      \multirow{ 2}{*}{GPT-3.5} & DL & 0.198 & 0.438 & 0.413  \\
       & DE & 0.229 & 0.461 & 0.500  \\
        & LE & 0.213 & 0.417 & 0.464  \\
                \midrule
    & DLE & 0.327 & 0.472 & 0.619  \\
      \multirow{ 2}{*}{GPT-4 Turbo} & DL & 0.280 & 0.458 & 0.560  \\
       & DE & 0.294 & 0.500 & 0.571  \\
        & LE & 0.299 & 0.514 & 0.583  \\

     \bottomrule
    \end{tabular}
    \caption{Argument reconstruction performance only over fallacies that could be detected based on linked passages (or based on the accurate premise alone).}
    \label{tab:subtask3-missci-onlymapped}
\end{table*} 

We report the argument reconstruction of the evaluated LLMs in Table~\ref{tab:subtask3-missci-onlymapped}. This evaluation only considers a prediction of an LLM  based on the paraphrased information $s_i$ if a passage $S_j$ that communicates the same information exists, and the same fallacy could have been detected based on $S_j$. This serves as a complementary comparison between the performance on \missci{} and \dataset{}, which removes benefits over \missci{} due to additional information.

\section{AFC applied on \dataset{}}

\subsection{Stance Predictions per Fallacy Class}
\label{appendix:afc-on-missci}

\begin{figure*}
\small
    \centering
    \includegraphics[width=\textwidth]{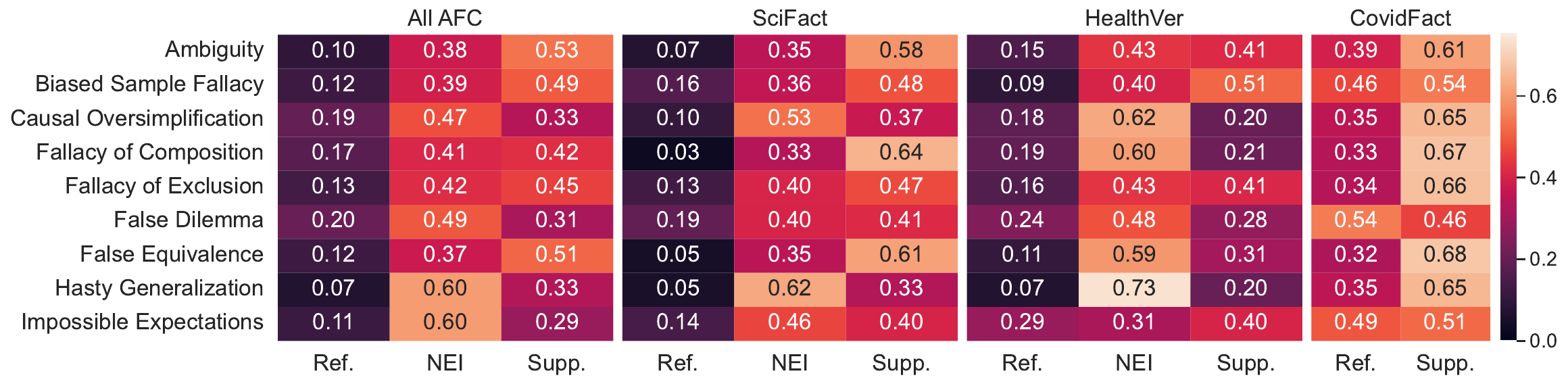}
    \caption{Distribution over predicted fact-checking labels per fallacy class, which was linked to the passage used as evidence, when predicting the veracity label. Results are averaged over five seeds for each fine-tuned AFC model.}  
    \label{fig:afc-classification-over-fallacies_heatmap}
\end{figure*}

We aim to understand how passages that point to different reasoning gaps (and hence different fallacy classes) affect the stance prediction of scientific AFC models. Figure~\ref{fig:afc-classification-over-fallacies_heatmap} visualizes for each AFC model and fallacy class the distribution over the predicted veracity labels of \textsc{Supported}, \textsc{Refuted} and \textsc{NEI}. Regardless of the fallacy class, rarely any passage is identified as refuting the claim. Most models vary between the labels \textsc{Supported} and \textsc{NEI} with only minor clear trends among different fallacy classes. Interestingly, the model trained on \textsc{CovidFact}, which had the best misinformation detection rate according to Table~\ref{tab:afc-on-missciplus} (as it does not know the label \textsc{NEI}) almost always tends to predict \textsc{Supported} more frequently than \textsc{Refuted} over these passages that exhibit fallacious reasoning behind the claim.

\subsection{Binary Fallacy Detection by LLMs}
\label{appendix:afc-binary-fallacy-detection}
\begin{table}[]
\small
    \centering
    \begin{tabular}{l | cc }
    \toprule
    & \multicolumn{2}{c}{\emph{Detected a fallacy}} \\
    \textbf{LLM} & \textbf{\dataset{}} & \textbf{True Claims} \\
    \toprule
       Llama2-70B & 0.988 & 0.993 \\ 
       Llama3-8B & 0.877 & 0.783 \\ 
       GPT-3.5 & 0.853 & 0.923 \\ 
     \bottomrule
     
    \end{tabular}
    \caption{Binary fallacy evaluation over misinformation from \dataset{} and correct claims from \textsc{HealthVer} and \textsc{CovidFact}.}
    \label{tab:binary-fallacy-true-missciplus}
\end{table} 

We adjusted our prompt (cf. §\ref{appendix:llm-as-afc-prompts}) to allow LLMs to select no fallacy if they consider a claim correct and empirically evaluate three LLMs in how well they detect misinformation in \dataset{} or correct information in the 100 selected claims from \textsc{HealthVer} and \textsc{CovidFact}. Since true claims only come with one evidence passage, we also prompt the LLMs once only for misinformation using the concatenated relevant passages (cf. \emph{all passages} in §\ref{sec:experiments:per-passage-prompting}).
Table~\ref{tab:binary-fallacy-true-missciplus} reports the ratio of claims identified as fallacious (the LLM did not specifically say that ``no fallacy'' exists \emph{and} found at least one fallacy) across both datasets averaged over three seeds. We do not discern \emph{which} fallacy was detected. Using the prompts to reconstruct fallacious arguments, the LLMs tend to find fallacies in all claims regardless of whether they are correct. We note that our claims are designed to correctly verbalize and identify all applied fallacies in \dataset{}. This follows previous works on fallacy detection \citep{da-san-martino-etal-2019-fine,jin-etal-2022-logical, alhindi-etal-2022-multitask} that focused on which fallacy was applied rather than determining \emph{if} a fallacy was applied. Future work may explore how to detect fallacies and correct claims better.

\subsection{LLMs may decline to provide a veracity prediction.}
The percentages of the label classification from LLMs as fact-checking models using their parametric knowledge or RAG evidence in Table~\ref{tab:llm-as-afc} may not sum up to 100\% if the LLM declines to answer. We provide an extended Table that includes the percentage when the LLM declines to answer in Table~\ref{tab:llm-as-afc-no-answer}.

\begin{table*}[]
\small
    \centering
    \begin{tabular}{ll|cccc|cccc}
    \toprule
    && \multicolumn{8}{c}{\emph{Predicted as}} \\
     & \textbf{LLM} & \textbf{True} & \textbf{False} & \textbf{NEI} & \textbf{No Answer} & \textbf{True} & \textbf{False} & \textbf{NEI} & \textbf{No Answer} \\
        \midrule
    & Llama2 &  1.6 & 61.1 & 37.3 & 0.0 & 34.7 & 22.3 & 41.3 & 1.7 \\
    Know & Llama3 & 0.0  & 86.9 & 2.4 & 10.7 & 20.0  & 43.3 & 14.3 & 22.3 \\
    (\emph{FC}) & GPT4 & 0.0  & 85.3 & 14.7 & 0.0 & 59.0 & 23.0 & 17.3 & 0.7\\
    & GPT3.5 & 0.8 & 71.0 & 28.2 & 0.0 & 46.7 & 17.3 & 35.7 & 0.3 \\
    \midrule
    & Llama2 & 0.0 & 100.0 & 0.0 & 0.0 & 29.7 & 69.3 & 1.0 & 0.0 \\
    Know & Llama3 & 8.3  & 88.9 & 2.4  & 0.4 & 68.7  & 26.0 & 3.0 & 2.3 \\
    (\emph{Ask}) & GPT4 & 3.6 & 68.3 & 27.8 & 0.4& 49.7 & 6.7 & 36.3 & 7.3\\
    & GPT3.5 & 1.6 & 50.4 & 48.0 &0.0& 47.3 & 6.0 & 45.0 & 1.7\\
    \midrule
        & Llama2 & 23.8 & 61.5 & 12.7 & 2.0 & 58.7 & 29.7 & 10.7 & 1.0\\
    \multirow{ 2}{*}{RAG} & Llama3 & 44.4  & 53.2 & 2.4 & 0.0 & 80.3  & 16.3 & 3.3 & 0.0 \\
    & GPT4 & 27.4 & 34.1 & 38.5 & 0.0 & 55.0 & 4.0 & 41.0 & 0.0 \\
   & GPT3.5 & 38.9 & 31.7 & 29.0 & 0.4 & 78.0 & 5.3 & 16.0 & 0.7\\

    \midrule

    \multicolumn{2}{c}{} & \multicolumn{4}{c|}{\textbf{Misinformation}} & \multicolumn{4}{c}{\textbf{True claims}} \\ 
    \bottomrule
    \end{tabular}
    \caption{Averaged veracity predictions from LLMs on misinformation from \dataset{} (\emph{left}) and true claims from \textsc{HealthVer} and \textsc{CovidFact} (\emph{right}).}
    \label{tab:llm-as-afc-no-answer}
\end{table*}

\section{Prompts}
\label{appendix:prompts}

\subsection{PRP Prompts}
\label{appendix:prp-prompts}
The prompt template for PRP ranking is shown in Figure~\ref{fig:prompt-prp} with the different \texttt{[INSTRUCTION]} listed in Table~\ref{tab:prp-instructions}.

    \begin{figure}[]
    \begin{tcolorbox}[left=1pt, right=1pt, top=1pt, bottom=1pt, boxrule=0.5pt]
    \small
\texttt{[INSTRUCTIONS]} \\

Passage A: \texttt{[PASSAGE-A]} \\
Passage B: \texttt{[PASSAGE-B]} \\ \\
Output Passage A or Passage B:

    \end{tcolorbox}
    \caption{Prompt template for PRP prompting.}
    \label{fig:prompt-prp}
    \end{figure}

    \begin{table*}[]
\small
    \centering
    \begin{tabularx}{\textwidth}{l|X}
    \toprule
    \textbf{Name} & \textbf{Instructions} \\
    \midrule
    \citet{qin2023large} & Given a query \texttt{[CLAIM]}, which of the following two passages is more relevant to the query? \\
    \citet{qin2023large} (claim)& Given a claim \texttt{[CLAIM]}, which of the following two passages is more relevant to the claim? \\
    \citet{qin2023large} (convincing)&  Given a claim  \texttt{[CLAIM]}, which of the following two passages constitutes more convincing evidence for the claim? \\
    \citet{qin2023large} (support)&  Given a claim \texttt{[CLAIM]}, is it more likely that someone made the claim based on passage A or based on passage B? \\
    
    \bottomrule
    \end{tabularx}
    \caption{Evaluated instructions used for PRP prompting.}
    \label{tab:prp-instructions}
\end{table*}

\subsection{Premise judge $\phi^{\mathrm{p+f}}$}
\label{appendix:judge-prompts}

    \begin{figure}[]
    \begin{tcolorbox}[left=1pt, right=1pt, top=1pt, bottom=1pt, boxrule=0.5pt]
    \small
\texttt{[INSTRUCTIONS]} \\\\
Premises:\\
1: "\texttt{[GEN-PREMISE]}" \\
2: "\texttt{[GOLD-PREMISE]}" \\ \\
Question: Do both premises use the same flawed reasoning (fallacy)?

    \end{tcolorbox}
    \caption{Prompt for the judge $\phi^{\mathrm{p+f}}$.}
    \label{fig:prompt-judge}
    \end{figure}

    \begin{table*}[]
\small
    \centering
    \begin{tabularx}{\textwidth}{l|X}
    \toprule
    \textbf{Name} & \textbf{Instructions} \\
    \midrule

   p1 & You are given two premises that exhibit some reasoning of a larger argument. Both premises apply a fallacy. Your task is to select whether the reasoning on an abstract level is identical. If one premise is more specific, they can still apply the same false reasoning. Provide your answer in the first line of your response. Answer with ``match'' if both premises apply the same false reasoning. Answer with ``no-match'' if they apply different false reasoning. \\
   \midrule
   p2 & I'll present you with two premises, each containing a fallacy in their reasoning. Analyze both statements: Your task: Determine if the core flawed logic behind the fallacies in both statements is identical. Respond with ``match'' if the underlying reasoning is the same, even if the specifics differ. Respond with ``no-match'' if they represent different fallacies. \\
   \midrule
   p3 & Task: Analyze both premises and determine if they commit the same type of fallacy. \\
    & Answer: (match / no-match)\\
   \midrule
   p4 &     Determine whether two premises exhibit identical false reasoning, regardless of their specificity.     Provide your answer in the first line of your response. Answer with ``match'' if both premises apply the same false reasoning. Answer with ``no-match'' if they apply different false reasoning.\\
   \midrule
   p5 &     Task: Analyze both premises and determine if they apply a similar reasoning regardless of specificity.\\
   &     Answer: (match / no-match)\\

    \bottomrule
    \end{tabularx}
    \caption{Instructions used for the judge $\phi^{\mathrm{p+f}}$.}
    \label{tab:judge-instructions}
\end{table*}

The prompt template for the judge $\phi^{\mathrm{p+f}}$ with ICL and SFT approaches is shown in Figure~\ref{fig:prompt-judge}. The instructions are listed in Table~\ref{tab:judge-instructions}.

\subsection{Argument reconstruction prompts}
\label{appendix:reconstruction-prompts}
    \begin{figure*}[]
    \begin{tcolorbox}[left=1pt, right=1pt, top=1pt, bottom=1pt, boxrule=0.5pt]
    \small
\texttt{[FALLACY INFORMATION]} \\

Task:\\
Examine the following fallacious argument: \\

Passage 1: "\texttt{[PASSAGE S0]}" \\
Passage 2: "\texttt{[PASSAGE Sj]}" \\
Claim: "\texttt{[CLAIM]}" \\

Passages 1 and 2 are sourced from the same credible scientific document. The claim is based on the content of both passages.
Your task is to identify and verbalize the fallacious reasoning as the fallacious premise necessary to support the claim given the content of passages 1 and 2.
This reasoning should effectively support the claim, ensuring that the passages do not undermine the claim as a valid conclusion.
Only consider fallacies from the provided fallacy inventory. \\

Present each fallacious premise along with the applied fallacy class in this format: \\
    \emph{Fallacious Premise: <fallacious premise>; Applied Fallacy Class: <applied fallacy class>}. \\
If multiple applicable fallacies exist, list all fallacious premises with the applied fallacy classes in order of relevance (most to least relevant).

    \end{tcolorbox}
    \caption{Prompt template for argument reconstruction \emph{p7-passage}.}
    \label{fig:prompt-per-passage-p7}
    \end{figure*}

    \begin{figure*}[]
    \begin{tcolorbox}[left=1pt, right=1pt, top=1pt, bottom=1pt, boxrule=0.5pt]
    \small
\texttt{[FALLACY INFORMATION]} \\

Task:\\
Examine the following fallacious argument: \\

Passage 1: "\texttt{[PASSAGE S0]}" \\
Passage 2: "\texttt{[PASSAGE Sj]}" \\
Claim: "\texttt{[CLAIM]}" \\

Passages 1 and 2 are sourced from the same credible scientific document. The claim is based on the content of both passages.
Your task is to identify and verbalize the fallacious reasoning (the hidden assumptions) as the fallacious premise necessary to support the claim given the content of passages 1 and 2.
This reasoning should effectively support the claim, ensuring that the passages do not undermine the claim as a valid conclusion.
Only consider fallacies from the provided fallacy inventory. \\

Present each fallacious premise along with the applied fallacy class in this format: \\
    \emph{Fallacious Premise: <fallacious premise>; Applied Fallacy Class: <applied fallacy class>}. \\
If multiple applicable fallacies exist, list all fallacious premises with the applied fallacy classes in order of relevance (most to least relevant).

    \end{tcolorbox}
    \caption{Prompt template for argument reconstruction \emph{p8-passage-assumptions}.}
    \label{fig:prompt-per-passage-p8}
    \end{figure*}

    \begin{figure*}[]
    \begin{tcolorbox}[left=1pt, right=1pt, top=1pt, bottom=1pt, boxrule=0.5pt]
    \small
\texttt{[FALLACY INFORMATION]} \\

\textbf{Challenge:} \\
    You've been assigned a detective mission in the world of scientific arguments! 

\textbf{The Case:} \\
    An argument has been made based on two passages from a credible scientific document. However, there's a hidden flaw in the reasoning. Your job is to uncover this hidden assumption – the "fallacious premise" – that makes the argument illogical.

\textbf{The Evidence:} \\
    Passage 1 (This is the first piece of information from the scientific document.): "\texttt{[PASSAGE S0]}"
    Passage 2 (This provides additional details from the same document.): "\texttt{[PASSAGE Sj]}"
    Claim (This is the conclusion drawn from both passages): "\texttt{[CLAIM]}"

\textbf{The Tools:} \\
    Fallacy Inventory: You have access to a list of common fallacies (errors in reasoning). Only use these fallacies!

\textbf{Your Mission:} \\
    1. Analyze the passages and the claim. \\
    2. Identify the hidden assumption that's needed for the claim to follow logically from the passages. \\
    3. Formulate this hidden assumption as a clear "fallacious premise." \\
    4. Identify the specific type of fallacy from the fallacy inventory that best explains this hidden assumption.

\textbf{Deliverables:} \\
    Present your findings in this format: \\
    \emph{"Fallacious Premise: <fallacious premise>; Applied Fallacy Class: <applied fallacy class>."} \\
    If multiple applicable fallacies exist, list all fallacious premises with the applied fallacy classes in order of relevance (most to least relevant).
    Only consider fallacies from the provided fallacy inventory.

\textbf{Remember:} \\
    The passages come from a credible scientific document, so the information itself is likely true.
    The fallacy lies in how the information is used to support the claim.
    Focus on the hidden assumption(s) needed to bridge the gap between the passages and the claim.

    \end{tcolorbox}
    \caption{Prompt template for argument reconstruction \emph{p9-passage-detective}.}
    \label{fig:prompt-per-passage-p9}
    \end{figure*}

    \begin{figure*}[]
    \begin{tcolorbox}[left=1pt, right=1pt, top=1pt, bottom=1pt, boxrule=0.5pt]
    \small
\texttt{[FALLACY INFORMATION]} \\

\textbf{Task:}\\
    This activity focuses on identifying weaknesses in scientific arguments based on source materials.

\textbf{Materials:}\\
    Passage 1: "\texttt{[PASSAGE S0]}" \\
    Passage 2: "\texttt{[PASSAGE Sj]}" \\
    Claim: "\texttt{[CLAIM]}" \\
    The claim is derived from the content of both passages.
    Both passages stem from the same credible scientific publication.

\textbf{Objective:}\\
    Analyze the relationship between the provided passages and the claim. Identify any assumptions or gaps in reasoning that might weaken the argument's validity.

\textbf{Instructions:}\\
    1. Read Passage 1, Passage 2, and the claim carefully. \\
    2. Consider how the information in the passages connects to the claim. \\
    3. Identify any missing information or hidden assumptions that would be necessary for the claim to logically follow from the passages. \\
    4. Formulate these missing pieces as clear "fallacious premises." \\
    5. Using the provided fallacy list, identify the type of fallacy associated with each fallacious premise.

\textbf{Output:}\\
    Present your findings in this format:\\
    \emph{"Fallacious Premise: <fallacious premise>; Applied Fallacy Class: <applied fallacy class>."}\\
    If multiple applicable fallacies exist, list them in order of relevance (most to least relevant).
    Only consider fallacies from the provided fallacy inventory.

\textbf{Important Note:}\\
    The passages are sourced from a credible scientific document, so the information itself is most likely accurate.
    The focus is on how the information is used to support the claim, not questioning the scientific content.
    \end{tcolorbox}
    \caption{Prompt template for argument reconstruction \emph{p10-passage-evaluate}.}
    \label{fig:prompt-per-passage-p10}
    \end{figure*}

    \begin{figure*}[]
    \begin{tcolorbox}[left=1pt, right=1pt, top=1pt, bottom=1pt, boxrule=0.5pt]
    \small
\texttt{[FALLACY INFORMATION]} \\

\textbf{Task:}\\
    This activity focuses on identifying weaknesses in scientific arguments based on source materials.

\textbf{Materials:}\\
    Passage 1: "\texttt{[PASSAGE S0]}" \\
    Passage 2: "\texttt{[PASSAGE Sj]}" \\
    Claim: "\texttt{[CLAIM]}" \\
    The claim is derived from the content of both passages.
    Both passages stem from the same credible scientific publication.

\textbf{Objective:}\\
    Analyze the relationship between the provided passages and the claim. Identify any assumptions or gaps in reasoning that might weaken the argument's validity.

\textbf{Instructions:}\\
    1. Read Passage 1, Passage 2, and the claim carefully. \\
    2. Consider how the information in the passages connects to the claim. \\
    3. Identify any missing information or hidden assumptions that would be necessary for the claim to logically follow from the passages. \\
    4. Formulate these missing pieces as clear "fallacious premises." \\
    5. Using the provided fallacy list, identify the type of fallacy associated with each fallacious premise.

\textbf{Output:}\\
    Present your findings in this format:
    "Fallacious Premise: <fallacious premise>; Applied Fallacy Class: <applied fallacy class>."
    If multiple applicable fallacies exist, list all fallacious premises with the applied fallacy classes in order of relevance (most to least relevant).
    Only consider fallacies from the provided fallacy inventory.

\textbf{Important Note:}\\
    The passages are sourced from a credible scientific document, so the information itself is most likely accurate.
    The focus is on how the information is used to support the claim, not questioning the scientific content.
    \end{tcolorbox}
    \caption{Prompt template for argument reconstruction \emph{p11-passage-evaluate-2}.}
    \label{fig:prompt-per-passage-p11}
    \end{figure*}

    \begin{figure*}[]
    \begin{tcolorbox}[left=1pt, right=1pt, top=1pt, bottom=1pt, boxrule=0.5pt]
    \small
\texttt{[FALLACY INFORMATION]} \\
\textbf{Challenge:} \\
    You've been assigned a detective mission in the world of scientific arguments!

\textbf{The Case:}\\
    An argument has been made based on two passages from a credible scientific document. However, there's a hidden flaw in the reasoning. Your job is to uncover this hidden assumption – the "fallacious premise" – that makes the argument illogical.

\textbf{The Evidence:}\\
    Passage 1 (This is the first piece of information from the scientific document.): "\texttt{[PASSAGE S0]}" \\
    Passage 2 (This provides additional details from the same document.): "\texttt{[PASSAGE Sj]}" \\
    Claim (This is the conclusion drawn from both passages): "\texttt{[CLAIM]}"

\textbf{The Tools:}\\
    Fallacy Inventory: You have access to a list of common fallacies (errors in reasoning). Only use these fallacies!

\textbf{Your Mission:} \\
    1. Analyze the passages and the claim. \\
    2. Identify the hidden assumption that's needed for the claim to follow logically from the passages. \\
    3. FOR EACH identified flaw: \\
        a. Formulate this hidden assumption as a clear "fallacious premise." \\
        b. Using the provided fallacy  inventory, identify the specific fallacy class from the provided fallacy inventory that best explains this fallacious premise. \\
        c. Output each identified flaw in a separate line following this format: "Fallacious Premise: <fallacious premise>; Applied Fallacy Class: <applied fallacy class>." 

\textbf{Deliverables:}\\
    Present your findings in this format: \\
    \emph{"Fallacious Premise: <fallacious premise>; Applied Fallacy Class: <applied fallacy class>." }\\
    If multiple applicable fallacies exist, list all fallacious premises with the applied fallacy classes in order of relevance (most to least relevant).
    Only consider fallacies from the provided fallacy inventory.

\textbf{Remember:}\\
    The passages come from a credible scientific document, so the information itself is likely true.
    The fallacy lies in how the information is used to support the claim.
    Focus on the hidden assumption(s) needed to bridge the gap between the passages and the claim.
    \end{tcolorbox}
    \caption{Prompt template for argument reconstruction \emph{p12-passage-detective-2}.}
    \label{fig:prompt-per-passage-p12}
    \end{figure*}

    \begin{figure*}[]
    \begin{tcolorbox}[left=1pt, right=1pt, top=1pt, bottom=1pt, boxrule=0.5pt]
    \small
\texttt{[FALLACY INFORMATION]} \\
\textbf{Task:}\\
    This activity focuses on identifying weaknesses in scientific arguments based on source materials.

\textbf{Materials:}\\
    Passage 1: "\texttt{[PASSAGE S0]}" \\
    Passage 2: "\texttt{[PASSAGE Sj]}" \\
    Claim: "\texttt{[CLAIM]}" \\
    The claim is derived from the content of both passages.
    Both passages stem from the same credible scientific publication.

\textbf{Objective:}\\
    Analyze the relationship between the provided passages and the claim. Identify any assumptions or gaps in reasoning that might weaken the argument's validity.

\textbf{Your Mission:}\\
    1. Analyze the passages and the claim. \\
    2. Identify the hidden assumption that's needed for the claim to follow logically from the passages. \\
    3. FOR EACH identified flaw: \\
        a. Formulate this hidden assumption as a clear "fallacious premise." \\
        b. Using the provided fallacy inventory, identify the specific fallacy class from the provided fallacy inventory that best explains this fallacious premise. \\
        c. Output each identified flaw in a separate line following this format: "Fallacious Premise: <fallacious premise>; Applied Fallacy Class: <applied fallacy class>."

\textbf{Output:}\\
    Present your findings in this format: \\
    \emph{"Fallacious Premise: <fallacious premise>; Applied Fallacy Class: <applied fallacy class>."} \\
    If multiple applicable fallacies exist, list all fallacious premises with the applied fallacy classes in order of relevance (most to least relevant).
    Only consider fallacies from the provided fallacy inventory.

\textbf{Important Note:}\\
    The passages are sourced from a credible scientific document, so the information itself is most likely accurate.
    The focus is on how the information is used to support the claim, not questioning the scientific content.
    \end{tcolorbox}
    \caption{Prompt template for argument reconstruction \emph{p13-passage-detective-3}.}
    \label{fig:prompt-per-passage-p13}
    \end{figure*}

We take the p1-p6 prompts from \missci{} as-is. For \emph{all-passages} prompting we concatenate all passages (except for one $S^0_j$ passage), treating them as \emph{publication context} in the \missci{} prompt, and treating the left-out $S^0_j$ as the accurate premise. 
The \emph{per-passage} prompt templates for p7-p13 are shown in Figures~\ref{fig:prompt-per-passage-p7}-\ref{fig:prompt-per-passage-p13}., where we always replace \texttt{[PASSAGE S0]} with a randomly sampled passage $S_j^0$ and \texttt{[PASSAGE Sj]} with a passage $S_j$ linked to a fallacy. For all, we present the fallacy information consisting of the fallacy definition, logical form and example from literature identically to \missci{} prompts. The \emph{all-passage} prompts are minimally edited from these prompts and have one dedicated field for all selected passages. All used prompts are provided within our repository.

\subsection{LLM as AFC prompts}
\label{appendix:llm-as-afc-prompts}

    \begin{figure}[]
    \begin{tcolorbox}[left=1pt, right=1pt, top=1pt, bottom=1pt, boxrule=0.5pt]
    \small

Write a fact-checking article about the claim that "\texttt{[CLAIM]}"\\

Conclude your fact-checking article with a verdict in a single line: \emph{"Verdict: (your verdict label)".}
Choose one of the following verdicts: True, Unknown, False.

    \end{tcolorbox}
    \caption{Prompt to generate a fact-checking article given no evidence.}
    \label{fig:prompt:llm-as-afc:fc}
    \end{figure}

    \begin{figure}[]
    \begin{tcolorbox}[left=1pt, right=1pt, top=1pt, bottom=1pt, boxrule=0.5pt]
    \small

    Claim: "\texttt{[CLAIM]}" \\

To the best of your knowledge, what is the veracity of the claim?
Provide a thorough explanation supporting your decision, select one of the answers (True, False, Unknown) and output the veracity of the claim in a single line: \emph{"Veracity: (your veracity label)"}.

    \end{tcolorbox}
    \caption{Prompt to ask for the veracity of the claim given no evidence.}
    \label{fig:prompt:llm-as-afc:bk}
    \end{figure}

        \begin{figure}[]
    \begin{tcolorbox}[left=1pt, right=1pt, top=1pt, bottom=1pt, boxrule=0.5pt]
    \small

Study: \\
\texttt{[EVIDENCE]} \\

Rate if the claim "\texttt{[CLAIM]}" is correct based on the study. 
Answer with \emph{"Verdict: (Correct / Incorrect / Not Enough Information)"}

    \end{tcolorbox}
    \caption{Prompt to ask for the veracity of the claim given evidence.}
    \label{fig:prompt:llm-as-afc:ev}
    \end{figure}

        \begin{figure*}[]
    \begin{tcolorbox}[left=1pt, right=1pt, top=1pt, bottom=1pt, boxrule=0.5pt]
    \small

\texttt{[FALLACY INFORMATION]} \\ 

\textbf{Challenge:}\\
    You've been assigned a detective mission in the world of scientific arguments!

\textbf{The Case:}\\
    An argument has been made based on two passages from a credible scientific document.  Your job is to assess the logical reasoning behind this argument.

\textbf{The Evidence:}\\
    Passages (The used information from the scientific document): "\texttt{[EVIDENCE]}"
    Claim (This is the conclusion drawn from the passages): "\texttt{[CLAIM]}"

\textbf{The Tools:}\\
    Fallacy Inventory: You have access to a list of common fallacies (errors in reasoning). Only use these fallacies!

\textbf{Your Mission:}\\
    1. Analyze the passages and the claim. \\
    2. Evaluate whether the claim logically follows from the information presented in the passages. \\
    3. IF you identify flaws in the reasoning, pinpoint all hidden assumptions that's needed for the claim to follow logically. \\
    4. FOR EACH identified flaw: \\
        a. Formulate this hidden assumption as a clear "fallacious premise." \\
        b. Using the provided fallacy inventory, identify the specific fallacy class from the provided fallacy inventory that best explains this fallacious premise. \\
        c. Output each identified flaw in a separate line following this format: "Fallacious Premise: <fallacious premise>; Applied Fallacy Class: <applied fallacy class>." \\
    5. IF the claim logically follows from the passages (no fallacy present), then simply output: "No Fallacy" 

\textbf{Deliverables:}\\
    If the argument is sound (if no fallacy or only minor fallacies exist), clearly state "No Fallacy" in the output. \\
    If a fallacy is present, present your findings in this format:
    "Fallacious Premise: <fallacious premise>; Applied Fallacy Class: <applied fallacy class>." \\
    If multiple applicable fallacies exist, list all fallacious premises with the applied fallacy classes in order of relevance (most to least relevant).
    Only consider fallacies from the provided fallacy inventory.

\textbf{Remember:}\\
    The passages come from a credible scientific document, so the information itself is likely true.
    The fallacy lies in how the information is used to support the claim.
    Focus on the hidden assumption(s) needed to bridge the gap between the passages and the claim.
    If the scientific document supports the claim, no critical fallacy is applied.

    \end{tcolorbox}
    \caption{Fallacy generation prompt that can predict ``no fallacy''.}
    \label{fig:prompt:llm-as-afc:fallacy}
    \end{figure*}

The prompts used to directly predict the veracity of claims using LLMs are shown in Figures~\ref{fig:prompt:llm-as-afc:fc}-\ref{fig:prompt:llm-as-afc:ev}. We replace \texttt{[EVIDENCE]} with the concatenated evidence passages.
The adapted fallacy generation prompt which allows to output that no fallacy exists, is shown in Figure~\ref{fig:prompt:llm-as-afc:fallacy}.

\section{Reproducibility}
All experiments with Llama2 or Llama3 used the instruction-tuned LLM and were performed on 80GB A100 GPUs. We always used the 70B and 8B models for Llama2 and Llama3, respectively. All experiments with LLMs are averaged over three seeds. The only exception is the experiments with GPT-4 Turbo in Table~\ref{tab:subtask3-missci-vs-us}, which we only ran once due to computational costs. 
We used the API version \texttt{2023-10-01-preview} 
for GPT-3.5 (model version: \texttt{0613})\footnote{We used \texttt{gpt-35-turbo-16k} in the \emph{all passages} prompting experiments and \texttt{gpt-35-turbo} in all other experiments},
GPT-4 (model version: \texttt{0613}) and
GPT-4 Turbo (model version: \texttt{1106-Preview}).
We used Grammarly\footnote{\url{https://app.grammarly.com/}} in writing this paper.

\end{document}